\documentclass[11pt,a4paper]{article}
\usepackage[utf8]{inputenc}
\usepackage[T1]{fontenc}
\usepackage{lmodern}      
\usepackage{textcomp}
\usepackage{microtype}
\usepackage[margin=1in]{geometry}
\usepackage{setspace}
\usepackage{parskip}      
\usepackage{amsmath,amssymb,amsfonts}
\usepackage{mathtools}
\usepackage{graphicx}
\usepackage{booktabs}
\usepackage{array}
\usepackage{multirow}
\usepackage{longtable}
\usepackage{caption}
\usepackage[numbers,square,sort&compress]{natbib}
\usepackage[dvipsnames]{xcolor}
\definecolor{neongreen}{HTML}{39FF14}
\usepackage[colorlinks=true,
            linkcolor=blue!60!black,
            citecolor=neongreen,
            urlcolor=blue!60!black,
            breaklinks=true]{hyperref}
\hypersetup{
  pdftitle={A Systematic Benchmark of Intraoperative Ultrasound-to-MR Synthesis for Brain Tumour Surgery},
  pdfauthor={Santiago Cepeda; Olga Esteban-Sinovas; Rosario Sarabia; Ignacio Arrese},
  pdfsubject={Medical image synthesis; intraoperative ultrasound; MRI; deep learning},
  pdfkeywords={ioUS-MR synthesis; ReMIND; GAN; transformer; diffusion; benchmark}
}
\title{\bfseries A Systematic Benchmark of Intraoperative Ultrasound-to-MR Synthesis for Brain Tumour Surgery}
\author{%
  Olga Esteban-Sinovas\,\textsuperscript{$\dagger$} \quad
  Santiago Cepeda\,\textsuperscript{$\dagger$}\thanks{Corresponding author: \texttt{scepedac@saludcastillayleon.es}} \quad
  Ignacio Arrese \quad
  Rosario Sarabia \\[0.6em]
  \normalsize Department of Neurosurgery, Neurovascular Unit \\
  \normalsize R\'io Hortega University Hospital, Valladolid, Spain \\[0.25em]
  \normalsize Specialized Group in Biomedical Imaging and Computational Analysis (GEIBAC) \\
  \normalsize Instituto de Investigaci\'on Biosanitaria de Valladolid (IBioVALL), Valladolid, Spain \\[0.5em]
  \normalsize \textsuperscript{$\dagger$}\textit{These authors contributed equally to this work.}
}
\date{}
\begin{document}
\maketitle
\begin{abstract}
Intraoperative ultrasound (ioUS) is a versatile, cost-effective modality in brain tumour surgery, but its interpretation is difficult: acquisition planes are non-standard, artefacts are modality-specific, and its appearance differs markedly from the preoperative MRI on which surgical-planning tools, segmentation models and the surgeon's experience rely. Synthesising MRI-like images from ioUS could let this MRI-based infrastructure be reused intraoperatively without an extra scan. Most prior work evaluates a single architecture in isolation; to our knowledge, no benchmark has spanned architectural paradigms, inference regimes and downstream-task endpoints under a common protocol. We address this gap on the public ReMIND data set (76 patients; 153 paired ioUS/T2w and 104 paired ioUS/FLAIR studies; 60/16 patient-level train/held-out split). Six generators (four GAN baselines: Pix2Pix, SwinPix2Pix, CycleGAN, CUT; the transformer-augmented ResViT; and the few-step diffusion model SynDiff) were each trained under four inference regimes (2D, 2.5D, 2D\,+\,3D-refinement, full-3D) and two targets (T2w only; T2w\,+\,FLAIR multi-task), yielding 48 experiments. Image-fidelity metrics (SSIM, PSNR, MAE, LPIPS) were complemented by an nnU-Net~v2 downstream segmentation evaluation (tumour and resection cavity) and by subgroup analyses by histological grade and reoperation. No architecture dominated every axis, and, critically, perceptual quality tracked downstream utility most closely (LPIPS, $r=-0.66$, $p<0.001$), whereas higher SSIM was associated with worse utility ($r=-0.64$, $p<0.001$); SynDiff-2.5D best preserved downstream segmentation ($U_{\text{Dice}}=0.55$). Perceptual and downstream-task metrics should therefore be reported alongside or in preference to global SSIM, and architecture choice conditioned on surgical phase, patient history and clinical objective.
\end{abstract}

\noindent\textbf{Keywords:} intraoperative ultrasound; MRI synthesis; brain tumour surgery; GAN; transformer; diffusion model; ReMIND; benchmark

\bigskip
\section{Introduction}
\label{sec:introduction}
Maximal safe resection is the surgical objective most strongly associated with prolonged survival and preserved neurological function in patients with intra-axial brain tumours~\citep{sanai_extent_2008,herveyjumper_maximizing_2016}. Achieving this objective relies on accurate intraoperative localisation of tumour margins, eloquent cortex and sub-cortical white-matter tracts, which in current practice is delivered by neuronavigation systems anchored to preoperative magnetic resonance imaging (MRI). MRI offers unrivalled soft-tissue contrast and multi-parametric characterisation of the lesion, and its spatial fidelity is sufficient for preoperative planning. Once the dura is opened, however, this fidelity degrades rapidly: cerebrospinal fluid drainage, gravity, tissue retraction, tumour debulking and progressive cavity collapse jointly produce non-rigid intra-operative deformations, collectively referred to as \textit{brain shift}, of up to two centimetres at the cortical surface and several millimetres at depth~\citep{machado_deformable_2019,xiao_evaluation_2020}. As surgery progresses, the preoperative MRI therefore becomes an increasingly mis-registered prior, and the anatomy displayed to the surgeon by the navigation system no longer reflects the resection field. Intraoperative MRI (iMRI) can in principle re-anchor navigation to current anatomy, but the cost of the infrastructure, the disruption to surgical workflow and the limited number of acquisitions feasible during a single procedure restrict its availability to a small number of specialised centres. Intraoperative ultrasound (ioUS) has consequently emerged as the most practical bedside alternative: it is inexpensive, radiation-free, repeatable without disturbing the procedure, and modern 3D probes provide volumetric reconstructions registered to the navigation reference frame~\citep{rai_brainshift_2025,shetty_navigated_2021,unsgaard_ability_2005} Despite these advantages, ioUS has not displaced MRI as the primary navigation modality. Speckle, acoustic shadowing, a restricted and probe-orientation-dependent field of view, operator variability and the strong appearance gap with respect to MRI complicate both clinical interpretation and computational use of ioUS, particularly for algorithms trained on MRI such as tumour segmentation or atlas-based parcellation pipelines~\citep{dorent_mhvae_2023,dorent_unified_2026}. Bridging this modality gap, so that the rich pre-operative MRI information can be reused intraoperatively, is therefore a central technical challenge of image-guided tumour neurosurgery.
\subsection{Related work}
\label{sec:related}
The MRI/ioUS modality gap has been approached through two complementary strategies: \textit{multimodal registration}, which estimates a spatial transformation that aligns the two modalities without changing their appearance, and \textit{cross-modal synthesis}, which removes the appearance gap upstream by translating one modality into the intensity domain of the other. The present work belongs to the synthesis line, although its design is informed by limitations identified in both, as discussed in Sections~\ref{sec:related-registration} and~\ref{sec:related-synthesis}.
\subsubsection{Multimodal MRI/ioUS registration}
\label{sec:related-registration}
Classical registration pipelines rely on hand-crafted similarity metrics designed to be robust to the MRI/US appearance mismatch, most notably the Linear Correlation of Linear Combination ($\mathrm{LC^2}$) of Wein~et~al.~\citep{hutchison_global_2013,fuerst_automatic_2014} and self-similarity descriptors such as the Modality Independent Neighbourhood Descriptor (MIND~\citep{heinrich_mind_2012}) and its Self-Similarity Context (SSC~\citep{heinrich_mind_2013}) extension, which underpin the deformable engines of ANTs, NiftyReg and \texttt{deedsBCV}~\citep{modat_fast_2010}. On the public RESECT~\citep{xiao_resect_2017} and BITE~\citep{mercier_online_2012} cohorts these methods have been refined to near-voxel target registration errors (TRE) of $\approx$\,1.5--2\,mm: cDRAMMS~\citep{machado_deformable_2019} reports a mean TRE of 2.08--2.28\,mm across three sites and 758 landmarks; the LC$^2$-based pipeline of ImFusion ranked first in the CuRIOUS\,2018 challenge with a mean TRE of 1.57\,mm~\citep{xiao_evaluation_2020}; and the more recent Learn2Reg\,2024 and ReMIND2Reg\,2024 benchmarks still place a classical entry, NiftyReg, at the top of the post-resection leaderboard (TRE $\approx$\,2.87\,mm)~\citep{hansen_learn2reg_2025,dorent_brain_2025}. Comparable behaviour is observed in neighbouring anatomies: Kuklisova-Murgasova et al.~\citep{kuklisova-murgasova_registration_2013} demonstrated that the synthesis of a pseudo-US volume followed by block-matching aligned 3D fetal neurosonography with reconstructed MRI to within inter-rater landmark variability, anticipating the synthesis-then-registration paradigm reconsidered in this work. The integration of deep learning into MRI/ioUS registration has been gradual and its empirical gains heterogeneous. End-to-end deep registration networks (ranging from supervised 3D U-Net displacement regressors~\citep{zeineldin_towards_2020} to weakly-supervised generative frameworks for prostate MRI/TRUS~\citep{azampour_multitask_2024} and unsupervised dual-discriminator GANs for brain shift~\citep{rahmani_d2bgan_2024}) report competitive or superior accuracy on their training cohorts, yet generalise less reliably across centres and acquisition protocols than the optimisation-based baselines they seek to replace. This trend is exemplified by the ReMIND2Reg\,2024 and 2025 challenges, in which the strongest learned submissions are hybrid pipelines that reintroduce MIND-SSC features and convex optimisation on top of learned components (MCBO~\citep{wang_unsupervised_nodate} and the winning MCPO of Wang~et~al.~\citep{wang_unsupervised_2026}), and in which the leading purely learned descriptor (CrossKEY~\citep{morozov_3d_2025}) is trained patient-specifically on synthetic ioUS volumes generated from each subject's pre-operative MRI. Across both classical and learned pipelines, the appearance gap recurs as the principal limiting factor, and synthesis is increasingly used to attenuate it, which underscores the value of the systematic ioUS-to-MRI synthesis study pursued here.
\subsubsection{Cross-modal MRI/ioUS synthesis}
\label{sec:related-synthesis}
Cross-modal synthesis addresses the appearance gap explicitly, by mapping one modality into the intensity domain of the other so that downstream algorithms can operate on a single, modality-consistent representation. Early proof-of-concept work targeted fetal imaging, where US data are comparatively abundant: Jiao et al.~\citep{jiao_anatomy_2019,jiao_self-supervised_2020} proposed an anatomy-aware, self-supervised US$\to$MRI synthesis network trained on a single-gestational-age cohort, demonstrating the feasibility of unpaired translation between modalities with markedly different physical contrast-formation mechanisms. Subsequent diffusion-based formulations have extended this paradigm to the fetal brain, such as the dual-DDPM correlation framework DDIC of Silverstein~et~al.~\citep{silverstein_translation_nodate}. The extension of these methods to intra-operative neuro-oncology presents substantially more stringent requirements: paired training data are scarce, pathological anatomy is heterogeneous, and downstream clinical tasks impose tighter tolerances on anatomical fidelity. Equally important, the existing brain-specific literature is markedly asymmetric with respect to the synthesis direction. The majority of published work targets the MRI$\to$ioUS direction, where synthetic ioUS is exploited as a training surrogate for MRI-trained downstream models, for example to produce patient-specific real-time ioUS segmentation networks~\citep{dorent_patient-specific_2024}, to pre-train tumour-segmentation models on diffusion-synthesised ioUS~\citep{li_brain_2026}, to learn cross-modal keypoint descriptors~\citep{rasheed_lm2dk_2024,morozov_3d_2025}, or to drive differentiable physics-based renderers~\citep{bertramo_diffus_2025} and conditional latent-diffusion translators~\citep{jiang_cross-modal_2025}. A complementary line synthesises ioUS for data augmentation rather than modality translation, such as the morphology-guided two-step latent-diffusion model of Lasala et al.~\citep{lasala_two-step_2026}, in which generated image--label pairs measurably improve downstream tumour segmentation. The opposite direction, ioUS$\to$MRI, has received considerably less attention, even though it is precisely the direction required to reuse pre-operative MRI information during surgery and to make MRI-trained tumour and parcellation pipelines operable on the ioUS volume. To our knowledge, BrainVoxGen~\citep{singh_brainvoxgen_2023} is the only published brain study that targets ioUS$\to$MRI as its primary objective; its 3D Pix2Pix baseline, evaluated on $\approx$\,18 patients, was explicitly positioned as a reproducible benchmark for subsequent work but did not reach clinically usable fidelity. The hierarchical multimodal variational autoencoders MHVAE~\citep{dorent_mhvae_2023} and MMHVAE~\citep{dorent_unified_2026} model both directions jointly and define the current methodological reference for the task on ReMIND, though their downstream evaluation focuses predominantly on the MRI$\to$ioUS branch.
\subsubsection{Generative paradigms in medical image translation}
\label{sec:related-paradigms}
Architecturally, brain MRI/ioUS synthesis has rapidly explored the three canonical families of modern generative modelling. Convolutional adversarial networks (Pix2Pix, CycleGAN and their contrastive variants such as CUT~\citep{isola_pix2pix_2017,zhu_cyclegan_2017,park_cut_2020}) remain the de facto baseline owing to their fast single-pass inference, maturity and well-understood behaviour under paired or unpaired supervision; the high-resolution conditional GAN of Wang~et~al.~\citep{wang_pix2pixhd_2018} and the least-squares adversarial objective of Mao~et~al.~\citep{mao_lsgan_2017} further improve the resolution and training stability of this family. Transformer-augmented generators such as ResViT~\citep{dalmaz_resvit_2022} were introduced to inject long-range context into the bottleneck of an otherwise convolutional encoder--decoder, and have achieved strong results on multi-contrast MRI translation. Diffusion models, initially prohibitive at clinical resolutions, have become competitive through few-step adversarial-diffusion formulations such as SynDiff~\citep{ozbey_syndiff_2023} and conditional latent-diffusion designs such as CCLD~\citep{jiang_cross-modal_2025}, which reports the strongest MRI$\to$ioUS fidelity on ReMIND among standard adversarial, transformer and diffusion baselines. In parallel, application-driven adaptations such as the coarse-to-fine CycleGAN-plus-SynthMorph pipeline of Wang et al.~\citep{wang_coarse_2025,hoffmann_synthmorph_2022}, the physics-based differentiable renderer DiffUS~\citep{bertramo_diffus_2025} and the multi-network prostate benchmark of Salmanpour et al.~\citep{salmanpour_influence_nodate} have explored synthesis as a component of larger surgical or diagnostic pipelines rather than as an end in itself.
\subsection{Gaps in the literature}
\label{sec:gaps}
Despite this rapid methodological progress, the comparative evidence base for ioUS$\to$MRI synthesis in brain tumour surgery remains fragmented in ways that limit its translational impact. We identify four recurring gaps. \textit{Architecture and inference regime are entangled.} Almost every published study introduces a single architecture and evaluates it on a single inference regime (typically slice-wise 2D or full-volume 3D) without a controlled comparison to alternative regimes, so that the contribution of the generator family cannot be separated from the contribution of how depth context is supplied to it. \textit{Image-quality metrics are weak proxies for clinical utility.} Evaluations rely overwhelmingly on global image-quality metrics, typically PSNR, SSIM~\citep{wang_ssim_2004} and the perceptual LPIPS~\citep{zhang_lpips_2018}. Yet a high score on these does not guarantee clinical value: synthetic images can exceed 0.85 SSIM while losing clinically meaningful texture~\citep{salmanpour_influence_nodate}. Which fidelity metric best predicts performance on the eventual clinical task has not been systematically established, and metric choice is rarely validated against a downstream endpoint; visually plausible synthetic images therefore do not guarantee preservation of task-relevant anatomical information. \textit{The post-resection phase is under-evaluated.} The few studies that do report downstream metrics typically restrict themselves to registration TRE on the pre-resection regime, leaving the post-resection cavity, which is precisely the surgical phase in which ioUS-guided navigation is most needed, essentially unexplored. As a result, it remains unclear which combination of generative paradigm, inference regime and training target should be preferred when the intended use of the synthetic MRI is to support an MRI-trained tumour or resection-cavity segmentation pipeline in the operating room. \textit{The ioUS$\to$MRI direction is empirically under-characterised.} Brain-specific MRI/ioUS synthesis has been dominated by the MRI$\to$ioUS direction, in which synthetic ioUS supports the training of MRI-trained downstream models. The reverse direction, which is the clinically relevant one because it brings the ioUS volume into a representation on which MRI-trained tumour and parcellation algorithms can operate without retraining, has been addressed by a single dedicated brain study~\citep{singh_brainvoxgen_2023} and as a secondary branch of bidirectional models~\citep{dorent_mhvae_2023,dorent_unified_2026}, with no systematic comparison across architectural families or inference regimes published to date.
\subsection{Contributions}
\label{sec:contributions}
In this work, we address these gaps through a systematic, controlled benchmark of ioUS$\to$MRI synthesis for brain tumour surgery on the public ReMIND cohort~\citep{juvekar_remind_2024}. Our objective is explicitly not to introduce a new architecture, but to provide a unified experimental scaffold in which the choice of architectural family, the choice of inference regime and the choice of evaluation criterion can be disentangled. Specifically, the contributions of this paper are:
\begin{itemize}
  \item \textbf{A unified ioUS$\to$MRI benchmark on ReMIND dataset.} We curate 153 paired ioUS / T2-weighted MRI studies (and 104 paired ioUS / FLAIR studies) from 76 subjects covering both the pre-resection and the post-resection phase, with subject-level train and held-out test splits that prevent intra-patient leakage by construction. To our knowledge, this is the first systematic benchmark dedicated to the ioUS$\to$MRI direction in brain tumour surgery.
  \item \textbf{Three generative paradigms under identical supervision.} We re-implement, under a single training and evaluation pipeline, representative architectures from the three dominant families of medical image synthesis: four adversarial baselines (Pix2Pix~\citep{isola_pix2pix_2017}, the Swin-augmented SwinPix2Pix~\citep{verdicchio_enhancing_2025}, CycleGAN~\citep{zhu_cyclegan_2017} and CUT~\citep{park_cut_2020}), a transformer-augmented residual generator (ResViT~\citep{dalmaz_resvit_2022}) and a few-step adversarial-diffusion model (SynDiff~\citep{ozbey_syndiff_2023}).
  \item \textbf{Volumetric adaptations of 2D-native architectures.} All six generators were originally formulated as slice-wise 2D models. We derive 2.5D channel-stacked, hybrid 2D\,+\,3D-refinement, and fully volumetric variants of each, including a 3D extension of ResViT's ART bottleneck in which its transformer branch is adapted to 3D windowed attention, and a fully 3D adversarial-diffusion variant of SynDiff with sliding-window inference and Hanning-window blending. These adaptations are not described in the original publications of the corresponding architectures and constitute a methodological contribution in their own right.
  \item \textbf{Four inference regimes treated as a controllable cross-family axis.} Every architecture is instantiated under axial-only 2D slice-wise synthesis, context-enhanced 2.5D synthesis, hybrid 2D synthesis with a 3D refinement head and full volumetric 3D synthesis, so that the contribution of depth context can be isolated from the contribution of the generator family.
  \item \textbf{Single-target and multi-task supervision.} Each of the resulting 24 configurations is trained under T2w-only supervision and under T2w\,+\,FLAIR multi-task supervision with homoscedastic uncertainty weighting~\citep{kendall_uncertainty_2018}, yielding 48 controlled experiments.
  \item \textbf{Clinically grounded evaluation.} Beyond the standard PSNR/SSIM/LPIPS triad~\citep{wang_ssim_2004,zhang_lpips_2018}, we evaluate every synthetic volume through two independently trained nnU-Net~\citep{isensee_nnunet_2021} segmentation models (Seg-T2 and Seg-FLAIR) using Dice, HD95 and the surgical-margin-aware NSD$_{2\mathrm{mm}}$~\citep{nikolov_clinically_2021,maierhein_metrics_2024}, separately on the pre-resection (tumour) and post-resection (cavity) phases.
\end{itemize}

\section{Materials and methods}
\label{sec:materials}
\subsection{Dataset and subject-level split}
All experiments were performed on the publicly available ReMIND cohort (Brain Resection Multimodal Imaging Database, Brigham and Women's Hospital, Boston, MA, USA), which provides intra-operative 3D ultrasound (ioUS) acquired during brain tumour resection together with the corresponding pre-operative and intra-operative magnetic resonance (MR) examinations~\citep{juvekar_remind_2024}. For each subject we retained two ioUS volumes, each paired with the contemporaneous MR examination of the same surgical phase: the pre-resection study (ioUS acquired with the dura intact, paired with the pre-operative MR) and the post-resection study (ioUS acquired once the surgeon judged the resection complete or required intra-operative MRI to assess residual tumour, paired with the intra-operative MR). The two phases were therefore treated as independent paired studies of the same patient. The pre-operative T2-weighted MR (T2w) was used as the primary synthesis target, and the FLAIR sequence, when available, was added as a secondary target for the multi-task experiments.
Cases were excluded primarily because of low ultrasound image quality (insufficient acoustic coupling, large shadowing artefacts, near-empty foreground or marked probe-induced distortion that precluded reliable training and evaluation). After this quality-driven selection, the working cohort consisted of 76 subjects with 153 paired ioUS / T2w studies (approximately 2 studies per subject) and 104 paired ioUS / FLAIR studies. The dataset was split at the subject level with a fixed random seed: 80\,\% of the subjects (60 patients, 122 paired studies) were assigned to the training set and 20\,\% (16 patients, 31 paired studies) were held out as the test set. Because the pre- and post-resection studies of the same patient were always kept in the same partition, intra-subject leakage was prevented by construction.
\subsection{Pre-processing pipeline}
The ReMIND volumes are distributed in DICOM format. Because all images were exported from a clinical neuro-navigation system, they were already expressed in a common physical reference frame; however, the ioUS and pre-operative MR volumes are only approximately co-registered. A unified pre-processing pipeline was therefore applied to produce the data tensors that feed every model. The pipeline consists of five steps.
\textbf{(1) Format conversion.} All DICOM series were converted to NIfTI format, preserving the original physical metadata and the common spatial reference inherited from the navigation system.
\textbf{(2) Rigid co-registration.} Each ioUS / MR pair was rigidly co-registered using ImFusion Suite (ImFusion GmbH, Munich, Germany; v2.42.2). Following Wein et al.~\citep{hutchison_global_2013}, the ioUS volume was used as the reference image and the pre-operative MR (T2w; and FLAIR when available) as the moving image, with the Linear Correlation of Linear Combination (LC$^2$) similarity metric driving the optimisation. The resulting transform was applied to the MR volumes.
\textbf{(3) Resampling.} The ioUS volume was resampled to the MR voxel grid using B-spline interpolation, ensuring that, for every subject, each ioUS voxel had a one-to-one correspondence with an MR voxel.
\textbf{(4) Field-of-view cropping.} A tight per-case bounding box was extracted from the ioUS foreground mask (intensity above 1\,\% of the volume maximum) and applied to the resampled ioUS, T2w and FLAIR volumes. Cropped volume sizes ranged from $84^3$ to $251^3$ voxels.
\textbf{(5) Intensity normalisation.} Let $M_{fg}$ denote the foreground voxel set of a volume, and let $p_2$ and $p_{98}$ be the 2nd and 98th percentiles of intensity over $M_{fg}$. Ultrasound voxels were robust-percentile clipped and rescaled to $[-1, 1]$:
\begin{equation}
\hat{u}_i = -1 + 2 \cdot \frac{\mathrm{clip}(u_i, p_2, p_{98}) - p_2}{p_{98} - p_2}, \quad u_i \in M_{fg}.
\end{equation}
T2w and FLAIR voxels were z-scored using the foreground mean $\mu_{fg}$ and standard deviation $\sigma_{fg}$, clipped at $\pm 3\sigma$ and rescaled to $[-1, 1]$:
\begin{equation}
\hat{m}_i = \frac{1}{3}\,\mathrm{clip}\!\left(\frac{m_i - \mu_{fg}}{\sigma_{fg}},\,-3,\,3\right), \quad i \in M_{fg}.
\end{equation}
\textbf{Resizing.} For all 2D and 2.5D experiments, cropped slices were resized in-plane to $256 \times 256$ at training time (depth left at native size for the 2.5D channel stack). Full-3D experiments used the native cropped resolution and 3D patches were extracted on the fly.
\textbf{Data augmentation.} Light augmentation was applied during training only. For 2D and 2.5D models we used random horizontal / vertical flips ($p = 0.5$), random in-plane rotations of $\pm 15^\circ$ ($p = 0.3$) and ioUS-only intensity perturbation (gamma $\in [0.8, 1.2]$ plus Gaussian noise $\sigma \in [0, 0.03]$, $p \approx 0.4$). For full-3D models we used independent flips along each axis ($p = 0.5$) and small $\pm 10^\circ$ rotations on a random axis pair ($p = 0.3$); rotations were skipped in SynDiff-3D for speed reasons related to trilinear backward propagation.
\subsection{The four inference regimes}
A central methodological choice of this benchmark is to treat the inference regime as a controllable cross-family axis. Every architecture was instantiated under four regimes that trade depth context, parameter sharing and inference cost in different ways (Table~\ref{tab:regimes}). ioUS volumes are not acquired along a fixed anatomical plane: the probe orientation is determined intra-operatively by the surgical approach. After co-registration to the pre-operative MR and resampling to the MR voxel grid (Section~2.2), every volume shares the navigation reference frame, and we process slices along a single fixed axis of that frame, by convention the axial axis. All experiments use the same axis to keep the comparison controlled; no multi-axis sampling or fusion was performed. Every regime was instantiated under T2w-only and T2w + FLAIR (multi-task) configurations, yielding 48 experiments across the six architectures.
\begin{table}[ht]
\centering
\caption{Inference models shared by every architecture in this benchmark.}
\label{tab:regimes}
\footnotesize
\setlength{\tabcolsep}{4pt}
\begin{tabular}{@{}p{2.0cm}p{5.4cm}p{7.2cm}@{}}
\toprule
\textbf{Model} & \textbf{Input} & \textbf{Inference} \\
\midrule
2D & Single axial slice ($256 \times 256 \times 1$). & Axial-only, slice-by-slice prediction; predicted slices are stacked along the depth axis to form the output volume. \\
\addlinespace
2.5D & Three adjacent axial slices stacked as input channels ($\ldots \times 3$); the centre slice is the prediction target. & Axial-only, slice-by-slice prediction with through-plane context provided by the input channel stack. \\
\addlinespace
2D + 3D-refine & A frozen 2D / 2.5D generator produces an axial prediction volume; a small 3D refinement head consumes the assembled prediction (concatenated with the ioUS for ResViT and SynDiff) and emits a residual correction. & Two-stage axial inference: (i) the 2D generator runs slice-by-slice; (ii) one sliding-window pass of the 3D head over the assembled volume. \\
\addlinespace
Full-3D & Asymmetric or cubic 3D patches ($64 \times 64 \times 32$ for the GAN baselines, $64^3$ for SynDiff, $96^3$ for ResViT on a 24\,GB GPU). & Sliding-window 3D inference with overlap (16, 16, 16)--(32, 32, 16) and Hanning-window blending on the overlap region. \\
\bottomrule
\end{tabular}
\end{table}
\subsection{Models}
\begin{figure}[ht]
\centering
\includegraphics[width=\textwidth]{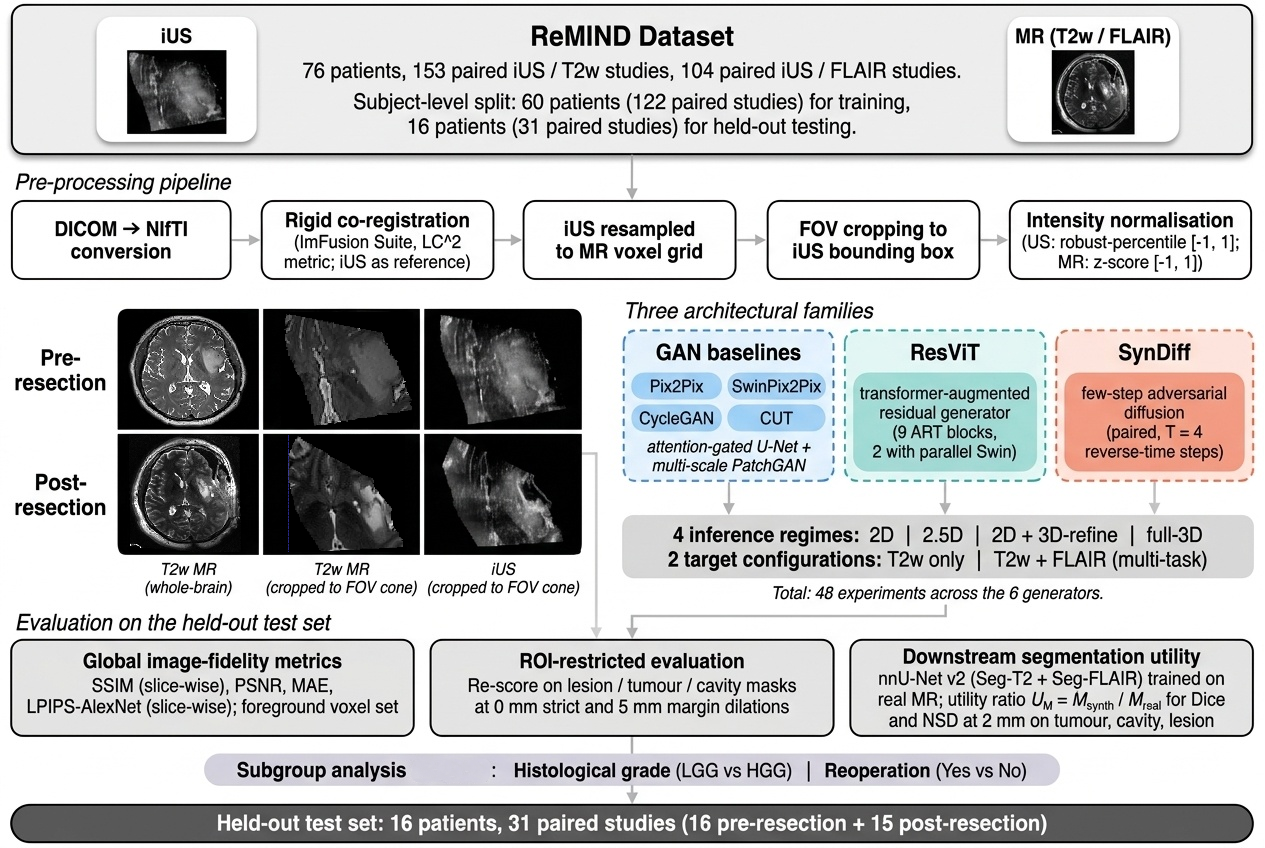}
\caption{Overview of the ioUS $\to$ MR synthesis benchmark. The pre-processing pipeline (DICOM $\to$ NIfTI conversion, ImFusion-LC$^2$ rigid co-registration, resampling of ioUS to the MR grid, FOV cropping to the ioUS volume and intensity normalisation) feeds three architectural families (GAN baselines, ResViT, SynDiff). Each architecture is instantiated under four inference regimes (2D, 2.5D, 2D + 3D-refine, full-3D) and two target configurations (T2w only and T2w + FLAIR multi-task), yielding 48 experiments evaluated on the held-out test set of 16 patients (31 paired studies).}
\label{fig:overview}
\end{figure}

\begin{figure}[!htbp]
\centering
\includegraphics[width=\textwidth]{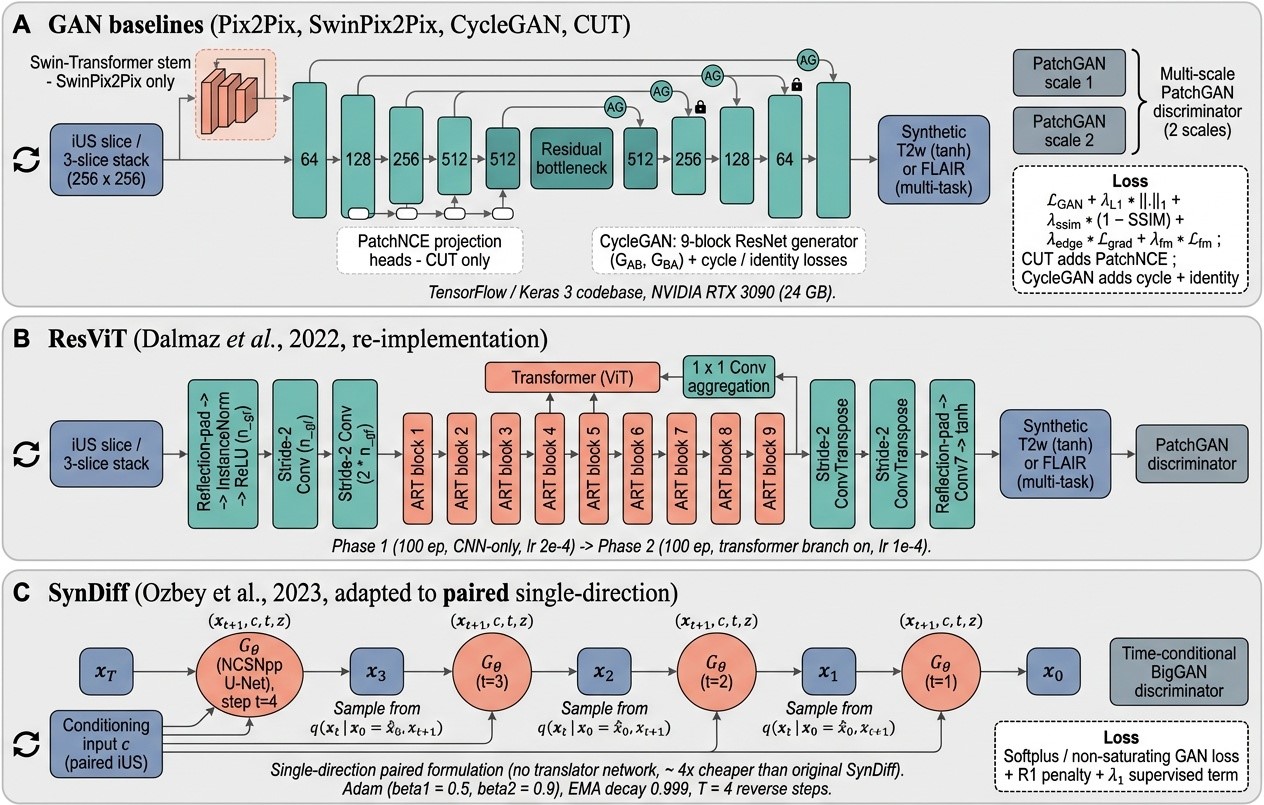}
\caption{Architectures of the six generators, shown in their 2D / 2.5D variant. \textbf{(A)} GAN baselines: Pix2Pix, SwinPix2Pix and CUT share an attention-gated U-Net generator (the Swin stem is used by SwinPix2Pix only, the PatchNCE heads by CUT only), while CycleGAN uses a 9-block ResNet generator; all use a multi-scale PatchGAN discriminator. \textbf{(B)} ResViT, with the transformer branch on ART blocks 4--5. \textbf{(C)} SynDiff in the single-direction paired formulation ($T=4$ steps). Training and loss details are given in Section~\ref{sec:materials}.}
\label{fig:architectures}
\end{figure}

Three architectural families were benchmarked under the four inference regimes of Section~2.3: classical GAN baselines (Section~2.4.1), a transformer-augmented residual generator (ResViT, Section~2.4.2), and an adapted few-step adversarial diffusion model (SynDiff, Section~2.4.3). Figure~\ref{fig:overview} summarises the overall data flow and Figure~\ref{fig:architectures} details the architectures and training schemes.

\subsubsection{Convolutional GAN baselines}
A unified TensorFlow / Keras codebase was developed to train the four GAN architectures listed below under identical hyper-parameters whenever possible. Every generator ends with a tanh activation that produces outputs in $[-1, 1]$.
\textbf{Pix2Pix.} Attention residual U-Net generator with five encoder blocks (filter widths 64--128--256--512--512; each block is a stride-2 Conv + LayerNorm + LeakyReLU(0.2) followed by a $3 \times 3$ residual block), a residual bottleneck and five decoder blocks with attention gates on each skip connection and dropout ($p = 0.5$) on the deepest skip. The final layer is a $3 \times 3$ Conv with tanh activation.
\textbf{SwinPix2Pix.} Same encoder / decoder skeleton as Pix2Pix, with a Swin-Transformer backbone embedded after the patch-embedding stem (patch size 2, embedding dimension 36, three stages with depths (2, 2, 2) and heads (6, 6, 12), window size 4, MLP ratio 4), shifted-window self-attention, and learnable relative-position bias.
\textbf{CycleGAN.} Standard 9-block ResNet generator with InstanceNorm~\citep{zhu_cyclegan_2017}, with the output layer adapted to the target channel count (1 for T2w-only, 2 for multi-task) and tanh activation. Both directions $G_{AB}$ (ioUS $\to$ MR) and $G_{BA}$ (MR $\to$ ioUS) are instantiated; cycle and identity losses are added on top of a paired-supervision L$_1$ + SSIM term that exploits the paired data.
\textbf{CUT (Contrastive Unpaired Translation).} The same attention residual U-Net as Pix2Pix is used as generator, but it additionally returns four encoder feature maps (initial conv, $e_1$, $e_2$, bottleneck). A per-layer MLP projection head maps these features into 128-dimensional $\ell_2$-normalised embeddings; the PatchNCE loss is computed on 256 randomly sampled spatial locations per layer with temperature $\tau = 0.07$. A paired $L_1 + \mathrm{SSIM} + \nabla$ term is added on top of the unpaired contrastive loss.
\textbf{Discriminators.} Multi-scale PatchGANs (two scales: original and 2$\times$ down-sampled) consisting of four spectrally-normalised $4 \times 4$ strided convolutions (64--128--256--512) with LayerNorm + LeakyReLU(0.2) and a final 1-channel logit map. Pix2Pix and SwinPix2Pix use conditional discriminators that receive the concatenation of (centre ioUS slice, target); CycleGAN and CUT use unconditional discriminators on the target only. Each discriminator returns its final logit map together with intermediate features for the feature-matching loss.
\textbf{Training objective.} For a paired sample $(x, y)$ and the generator prediction $\hat{y} = G(x)$, the discriminator and generator were optimised with least-squares GAN losses:
\begin{equation}
L_D = \tfrac{1}{2}\,\mathbb{E}\!\left[(D(x, y) - 1)^2\right] + \tfrac{1}{2}\,\mathbb{E}\!\left[D(x, G(x))^2\right], \quad
L_G^{adv} = \tfrac{1}{2}\,\mathbb{E}\!\left[(D(x, G(x)) - 1)^2\right].
\end{equation}
The generator was regularised by a multi-term supervised objective:
\begin{equation}
L_G = L_G^{adv} + \lambda_{L1}\,\|y - \hat{y}\|_1 + \lambda_{ssim}\,(1 - \mathrm{SSIM}(y, \hat{y})) + \lambda_{edge}\,L_\nabla + \lambda_{fm}\,L_{fm}.
\end{equation}
Here $L_\nabla$ is an edge / gradient $L_1$ term computed on first-order finite differences along the three image axes, and $L_{fm}$ is the standard feature-matching term on intermediate discriminator activations $\varphi_D$~\citep{wang_pix2pixhd_2018}. Discriminator stability was enforced by an $R_1$ gradient penalty applied lazily every 16 steps. CycleGAN additionally imposed cycle-consistency and identity losses; CUT replaced the unpaired adversarial signal with the per-layer PatchNCE objective:
\begin{equation}
L_{NCE}^{(\ell)} = -\sum_{s \in S} \log \frac{\exp(\langle q_s, k_s^+ \rangle / \tau)}{\exp(\langle q_s, k_s^+ \rangle / \tau) + \sum_{n \in N_s} \exp(\langle q_s, k_n^- \rangle / \tau)},
\end{equation}
where $q_s$ and $k_s^+$ are the query and positive embeddings at the same spatial location, $|N_s| = 255$ negative locations are sampled in the same image, and $\tau = 0.07$.
\textbf{Multi-task supervision.} In the T2w + FLAIR configuration the generator outputs two channels; per-channel $L_1 + \mathrm{SSIM} + \nabla$ losses are combined via the homoscedastic uncertainty weighting of Kendall et al.~\citep{kendall_uncertainty_2018}:
\begin{equation}
L_{MT} = \frac{1}{2\sigma_{T2}^2}\,L_{T2} + \frac{1}{2\sigma_{F}^2}\,L_F + \log \sigma_{T2} + \log \sigma_F.
\end{equation}
The two log-variance scalars $\log \sigma_{T2}^2$ and $\log \sigma_F^2$ were added to the generator's trainable parameters. CycleGAN and CUT skipped the identity term in 2.5D and multi-task settings because of channel mismatches between $G_{AB}$ and $G_{BA}$.
\textbf{Loss weights.} $\lambda_{L1} = 10$, $\lambda_{ssim} = 8$, $\lambda_{edge} = 5$, $\lambda_{fm} = 2$, $\gamma_{R1} = 10$ (every 16 steps), $\lambda_{cyc} = 10$, $\lambda_{idt} = 5$ (when applicable), $\lambda_{NCE} = 1$, $\lambda_{idt\text{-}NCE} = 0.5$ (when applicable); supervised $L_1 + \mathrm{SSIM} + \nabla$ weights for CycleGAN / CUT were 10, 8, 2.
\textbf{Common training settings.} Adam optimiser ($\beta_1 = 0.0$, $\beta_2 = 0.999$), gradient clip-norm 1.0, cosine learning-rate schedule with 1\,000-step linear warm-up, base generator learning rate $2 \times 10^{-4}$ (CycleGAN / CUT discriminator $2 \times 10^{-4}$; Pix2Pix / SwinPix2Pix discriminator $5 \times 10^{-5}$), and a Polyak / EMA copy of the generator with decay 0.999 used for evaluation. Training length was 20\,000 steps for 2D / 2.5D and 50\,000 steps for full-3D, plus 10\,000 additional steps of 3D-refiner training on top of cached 2D predictions for the 2D + 3D-refine variants. Batch sizes were 16 (Pix2Pix / SwinPix2Pix 2D / 2.5D), 8 (CUT), 4 (CycleGAN) and 1 (full-3D, $64 \times 64 \times 32$ patches). Mixed precision was disabled. Final inference was performed with the EMA weights.
\subsubsection{ResViT}
We re-implemented ResViT~\citep{dalmaz_resvit_2022}. The generator follows the original recipe.
\textbf{Encoder.} ReflectionPad--Conv7--IN--ReLU ($n_{gf}$ filters) followed by two stride-2 Conv3--IN--ReLU blocks ($2\,n_{gf}$ and $4\,n_{gf}$ filters); the bottleneck width is $b_n = 4\,n_{gf}$, with $n_{gf} = 64$ for 2D / 2.5D / 2D + 3D-refine and $n_{gf} = 24$ / $32$ / $48$ for full-3D depending on the VRAM budget.
\textbf{Bottleneck.} A stack of nine Aggregated Residual Transformer (ART) blocks. Blocks at positions (4, 5) carry a parallel transformer branch whose output is concatenated with the CNN branch and aggregated by a $1 \times 1$ Conv--IN--ReLU; the remaining seven blocks are CNN-only residual blocks. Each ART block has independent transformer weights (no sharing).
\textbf{Transformer branch.} In the 2D, 2.5D and 2D\,+\,3D-refine regimes the transformer branch is the original ResViT module, a vision-transformer block with global multi-head self-attention (eight heads, MLP ratio 4) and learnable position encoding. For the full-3D regime, dense global attention over the volume is intractable, so we replace it with a windowed (Swin-style) 3D self-attention using shifted windows of size 4; this windowed 3D attention is the only departure from the original ResViT design.
\textbf{Decoder and discriminator.} The decoder uses two ConvTranspose stride-2 stages followed by ReflectionPad--Conv7--Tanh. The discriminator is a PatchGAN with three stride-2 layers (64--128--256--512 channels), $4 \times 4$ convs, LeakyReLU and a 1-channel output. For the 2D + 3D-refine variant we appended a small 3D ResNet head (three IN--ReLU--Conv3 blocks of 32 channels and a tanh output) that learned a residual correction $0.1 \cdot R(x)$ on top of the cached 2D ResViT prediction.
\textbf{Losses.} LSGAN adversarial term ($\lambda_{adv} = 1.0$), strong $L_1$ ($\lambda_{L1} = 100$) and feature matching across all $D$ layers ($\lambda_{fm} = 10$). When targets were partially missing in the multi-task case, a masked $L_1$ normalised by the number of available channels times spatial voxels was used in place of the standard $L_1$.
\textbf{Two-phase training.} Phase 1 trained the CNN-only generator for 100 epochs at $\mathrm{lr} = 2 \times 10^{-4}$ with the transformer branch off. Phase 2 loaded the best Phase-1 checkpoint, enabled the transformer branch and continued for another 100 epochs at $\mathrm{lr} = 1 \times 10^{-4}$. Optimiser was Adam ($\beta_1 = 0.5$, $\beta_2 = 0.999$) with cosine annealing to $1 \times 10^{-6}$, gradient checkpointing on every ART block in the 3D variant, and gradient accumulation ($\mathrm{acc} = 4$) when batch size was 1 (3D). For the 2D + 3D-refine variant the refinement head was trained for 30 additional epochs at $\mathrm{lr} = 1 \times 10^{-4}$ on cached predictions.
\textbf{Inference.} Axial-only at test time: the 2D / 2.5D generator runs slice-by-slice and slices are stacked; the 2D + 3D-refine variant adds a single sliding-window pass of the refinement head over the assembled volume; the full-3D variant uses sliding-window inference with Hanning-window blending. No multi-axis fusion was used.
\subsubsection{SynDiff}
We adapted SynDiff~\citep{ozbey_syndiff_2023}, an adversarial diffusion model built on the DDGAN framework~\citep{xiao_ddgan_2022}, to the paired ioUS $\to$ T2w setting. The original bidirectional + cycle-consistent setup was simplified to a single-direction paired formulation: only the diffusive generator and its discriminator were retained, conditioned directly on the real paired ioUS instead of on a translator stand-in. This reduces training cost by approximately $4\times$ (two networks instead of eight) while exploiting the paired supervision.
\textbf{Generator (NCSNpp / NCSNpp3D).} Time-conditional U-Net with adaptive group-norm and Fourier-feature time embedding. Channel multiplier $[1, 1, 2, 2, 4, 4]$ over six resolutions, two residual blocks per resolution, BigGAN-style residual blocks, FIR up- / down-sampling kernel $[1, 3, 3, 1]$, attention at resolution 16 (2D) and at the bottleneck (3D), dropout 0, base width $n_{gf} = 32$ (2D and 2.5D) and $n_{gf} = 24$ (3D).
\textbf{Discriminator.} Five-stage down-sampling discriminator with time-conditional embedding ($t_{\mathrm{emb}\_\mathrm{dim}} = 256$); it takes $(x_t, x_{t+1})$ for real pairs and $(x_{pos}, x_{t+1})$ for fake pairs over two channels.
\textbf{Diffusion process.} A variance-preserving SDE was used:
\begin{equation}
\beta(t) = \beta_{\min} + t\,(\beta_{\max} - \beta_{\min}), \quad
\bar{\alpha}_t = \exp\!\left(-\int_0^t \beta(s)\,ds\right), \quad
x_t = \sqrt{\bar{\alpha}_t}\, x_0 + \sqrt{1 - \bar{\alpha}_t}\,\varepsilon,
\end{equation}
with $\beta_{\min} = 0.1$, $\beta_{\max} = 20$ and only $T = 4$ reverse-time steps, the DDGAN-style few-step regime. At each step, the generator $G_\theta$ receives $(x_{t+1}, c, t, z)$, where $c$ is the conditioning (paired ioUS) and $z \sim \mathcal{N}(0, I)$ is an $n_z = 100$-dimensional latent, and predicts $\hat{x}_0 = G_\theta(x_{t+1}, c, t, z)$. The next-step sample is drawn from the analytical posterior $q(x_t \mid x_0 = \hat{x}_0, x_{t+1})$.
\textbf{Adversarial diffusion losses.} Let $\sigma(\cdot)$ denote the logistic sigmoid. The discriminator is trained per timestep on $(x_t, x_{t+1})$ versus $(x_{pos}, x_{t+1})$ via a softplus / non-saturating GAN loss with $R_1$ penalty:
\begin{equation}
L_D = \mathbb{E}\!\left[-\log \sigma(D(x_t, t, x_{t+1}))\right] + \mathbb{E}\!\left[-\log(1 - \sigma(D(x_{pos}, t, x_{t+1})))\right] + \frac{\gamma_{R1}}{2}\,\|\nabla_{x_t} D\|_2^2.
\end{equation}
The generator is trained against the discriminator with an additional supervised $L_1$ on the predicted $\hat{x}_0$:
\begin{equation}
L_G = -\mathbb{E}\!\left[\log \sigma(D(\hat{x}_{pos}, t, x_{t+1}))\right] + \lambda_{L1}\,\|x_0 - \hat{x}_0\|_1.
\end{equation}
Hyper-parameters: $\gamma_{R1} = 1$ (applied lazily every 64 steps in 2D / 2.5D and every 32 steps in 3D), $\lambda_{L1} = 10$, $\mathrm{lr}_G = 1.6 \times 10^{-4}$, $\mathrm{lr}_D = 1 \times 10^{-4}$, Adam ($\beta_1 = 0.5$, $\beta_2 = 0.9$), cosine annealing to $1 \times 10^{-5}$ and EMA decay 0.999 on the generator. All SynDiff runs used bf16 autocast. Training length: 300 epochs (2D / 2.5D), 200 epochs (3D, batch 1, $\mathrm{samples\_per\_volume} = 4$), and 100 epochs of 3D-refinement on top of the frozen 2D paired SynDiff (a small 3D ResNet refiner of six ResBlock3D blocks at $n_{gf} = 24$, Conv7--tanh exit, residual output) trained adversarially against a 4-layer 3D PatchGAN.
\subsection{Training infrastructure}
All experiments were run on a single workstation equipped with an Intel Core i9 CPU, 64\,GB of system RAM and an NVIDIA GPU RTX 3090 (24\,GB VRAM) used for the PyTorch pipelines (ResViT, SynDiff and the nnU-Net segmentation models) and the TensorFlow / Keras 3 GAN-baseline pipeline. The PyTorch stack was based on PyTorch 2.5.1 with CUDA 12.1 and cuDNN 9; the TensorFlow stack was based on TensorFlow 2.20.0 with cuDNN 9.
\subsection{Evaluation protocol}
All models were evaluated on the same 16-patient held-out test set (31 paired studies). For every test study, a full prediction volume was reconstructed (no patch-level metrics) by running the model in its native axial-only inference mode: slice-by-slice 2D / 2.5D for the GAN baselines and ResViT; 4-step adversarial-diffusion sampling per slice for SynDiff 2D / 2.5D; sliding-window 3D for the full-3D variants; and the corresponding two-stage 2D-then-refinement pipeline for the 2D + 3D-refine variants. Predictions were written to NIfTI on the original FOV-cropped grid for downstream re-use.
Image-quality metrics were computed on the foreground voxel set $\Omega$, defined from the target intensity in the $[0, 1]$ space as $\Omega = \{\, i : y_i^{[0,1]} > 0.025 \,\}$, with the additional requirement that an axial slice contributed to the SSIM and LPIPS averages only if at least 1\,\% of its voxels belonged to $\Omega$. Four metrics were reported.
\textbf{Structural Similarity (SSIM).} Computed slice-wise along the axial axis with $\mathrm{data\_range} = 1.0$ and averaged over foreground slices:
\begin{equation}
\mathrm{SSIM}(x, y) = \frac{(2\mu_x\mu_y + c_1)(2\sigma_{xy} + c_2)}{(\mu_x^2 + \mu_y^2 + c_1)(\sigma_x^2 + \sigma_y^2 + c_2)}.
\end{equation}
\textbf{Peak Signal-to-Noise Ratio (PSNR).} Computed globally on foreground voxels with both volumes renormalised to $[0, 1]$:
\begin{equation}
\mathrm{PSNR} = 10\,\log_{10}\frac{1}{\mathrm{MSE}_\Omega}, \quad
\mathrm{MSE}_\Omega = \frac{1}{|\Omega|}\sum_{i \in \Omega}(y_i - \hat{y}_i)^2.
\end{equation}
\textbf{Mean Absolute Error (MAE).} Also taken over foreground voxels in the $[0, 1]$ space:
\begin{equation}
\mathrm{MAE} = \frac{1}{|\Omega|}\sum_{i \in \Omega}|y_i - \hat{y}_i|.
\end{equation}
\textbf{Learned Perceptual Image Patch Similarity (LPIPS).} Computed slice-wise along the axial axis using the AlexNet backbone of Zhang et al.~\citep{zhang_lpips_2018}; grayscale slices were replicated to three channels and rescaled to $[-1, 1]$. LPIPS was computed as the spatial average over channel-normalised AlexNet feature differences weighted by the LPIPS calibration weights, as defined in the original paper.
Per-experiment summary tables report the mean, the standard deviation and a 95\,\% Student-$t$ confidence interval on per-subject scores. Paired statistical comparisons across models used the Wilcoxon signed-rank test. All evaluation was axial-only and no cross-family ensembles were considered.
\subsubsection{ROI-restricted synthesis evaluation}
The image-quality metrics of Section~2.6 are computed over a broad foreground mask defined from the target intensity, which is dominated by anatomically homogeneous brain background. Because all 48 methods reproduce that background trivially, the global metrics partly reward off-target fidelity and compress between-method differences in the surgical region that ultimately matters for downstream interpretation. To stress the comparison around the lesion, we re-evaluated every method on three anatomically meaningful masks derived from the curated MRI segmentations used in Section~2.7: a lesion mask (label $\geq 1$; tumour $\cup$ resection cavity), a tumour-only mask (label $= 1$) and a cavity-only mask (label $= 2$). Each mask was evaluated at two dilations, 0\,mm (strict; the exact label boundary as drawn) and 5\,mm (margin; captures peritumoural infiltration and absorbs sub-voxel registration error between prediction and target). Dilations were computed with an anisotropy-aware Euclidean distance transform that respects the per-axis voxel spacing of the common FOV-cropped grid shared by all predictions ($A \times B \times C$\,mm); when needed, the segmentation was resampled to that grid by nearest-neighbour interpolation. The same four metrics of Section~2.6 (SSIM, PSNR, MAE, LPIPS) were re-computed with one critical change in the slice-based primitives: each axial slice was cropped to the ROI bounding box (padding 2\,px, minimum side 11\,px) before SSIM and LPIPS so that the large identical zero-background outside the lesion did not saturate similarity near 1 and erase between-method differences. Voxel-based metrics (PSNR, MAE) were averaged over ROI voxels only, with a minimum of 100 voxels per subject; slice-based metrics (SSIM, LPIPS) skipped slices containing fewer than 50 ROI pixels. Aggregation, leaderboards and paired Wilcoxon tests were carried out identically to Section~2.6 along the (ROI, dilation, phase) axes. The ROI-restricted re-score was applied to all 48 single-target and multi-task experiments and run on a cohort of $n = 30$ T2w-only studies and $n = 20$ multi-task (T2w + FLAIR) studies with available MRI segmentation.
\subsection{Downstream segmentation evaluation}
\label{sec:downstream}
The image-quality metrics of Section~2.6 measure the per-voxel fidelity of the synthetic MRI but do not address whether the synthetic image preserves enough anatomical information to be useful for downstream tasks trained on real MRI. We therefore complemented the synthesis benchmark with a paired downstream-task evaluation in which two independent segmentation models, trained respectively on real T2w and real FLAIR, were applied to every synthesis output and compared with their performance on the corresponding real MRI. By holding the segmentation model fixed and varying only the input modality, any performance difference can be attributed to the synthesis itself.
\textbf{Reference segmentations.} The downstream evaluation used the same subject-level split as the synthesis benchmark of Section~2.1 (60 training subjects---122 paired studies; 16 test subjects---31 paired studies). Reference labels were generated semi-automatically inside ITK-SNAP using the nnInteractive plug-in, and were subsequently reviewed and refined by an expert observer (a board-certified neurosurgeon). Two foreground classes were annotated, tumour and resection cavity, together with a background class. Annotations were performed directly on the pre-processed T2w and FLAIR volumes, that is, after co-registration to the ioUS reference and cropping to the ioUS field of view (Section~2.2), so that real and synthetic MRI inputs share an identical spatial reference.
\textbf{Segmentation models.} Two independent nnU-Net v2~\citep{isensee_nnunet_2021} segmentation models were trained in 3d\_fullres configuration: Seg-T2 (on real T2w) and Seg-FLAIR (on real FLAIR). The default PlainConvUNet topology was used, with five encoder / decoder stages (32--64--128--256--320 features), 3D instance normalisation and LeakyReLU activations, trained on patches of $48 \times 112 \times 112$ voxels with batch size 5 (T2w) or 3 (FLAIR). Optimisation used SGD with Nesterov momentum, initial learning rate 0.01 and a poly schedule, with a Dice + cross-entropy loss under deep supervision. Augmentation followed the default heavy 3D nnU-Net pipeline (random rotation, scaling, mirroring, gamma and contrast jitter, Gaussian noise and blur, low-resolution simulation). Both segmentation models were trained for 500 epochs, with the best-validation checkpoint retained per fold. A 5-fold subject-stratified cross-validation was performed within the training cohort, and all five folds were retained for ensemble inference.
\textbf{Normalisation of synthetic inputs.} Synthetic predictions were saved on the FOV-cropped grid in their native value range (broadly $[-1, 1]$ for tanh outputs and $[0, 1]$ for diffusion outputs). The nnU-Net internal per-volume foreground-mean z-score normalisation, declared as modality ``MRI'' in the training configuration, was applied identically to real and synthetic inputs at inference, harmonising their intensity distributions without manual rescaling.
\textbf{Inference.} For every test study, all candidate T2w inputs (one real T2w plus the synthetic predictions from the single-target GAN, ResViT and SynDiff configurations of Tables~\ref{tab:t2w_single} and~\ref{tab:t2w_multi}) were independently segmented by the Seg-T2 ensemble. The FLAIR analysis used the candidate inputs consisting of real FLAIR plus the FLAIR channel of each multi-task synthesis. Inference combined the five-fold soft argmax with sliding-window prediction ($\mathrm{tile\_step\_size} = 0.5$) and mirror test-time augmentation.
\textbf{Segmentation metrics.} For each (set, study, class $\in \{\text{tumour, cavity}\}$) we computed the Dice similarity coefficient, the 95th-percentile bidirectional surface-to-surface Hausdorff distance (HD95, in millimetres, computed on anisotropy-aware Euclidean distance transforms) and the Normalised Surface Dice at 2\,mm tolerance (NSD$_{2\mathrm{mm}}$; \citep{nikolov_clinically_2021}), which recent consensus guidance~\citep{maierhein_metrics_2024} recommends as more clinically relevant than Dice for tasks where surgical margins matter. Subject-level aggregation averaged the per-class scores across only those classes for which a ground-truth structure was present, so that the model was not penalised when a label was anatomically absent.
\textbf{Lesion class (tumour $\cup$ cavity).} In addition to the per-class metrics, we report a binary lesion class defined as the union of labels $\{\text{tumour, cavity}\}$ on both the ground-truth segmentation and the model prediction, treating the two foreground sub-components as a single class. This compound endpoint answers the operational question of whether the model localises the diseased region at all, irrespective of whether it labels the tumour / cavity boundary correctly; it is particularly informative for the post-resection subset, where confusion between residual tumour and the developing cavity at their shared interface can depress per-class Dice without reflecting a localisation failure. Dice, HD95 and NSD$_{2\mathrm{mm}}$ are computed identically to the per-class case, and lesion results are reported alongside tumour and cavity in every downstream aggregation, Wilcoxon comparison and dual cross-modal table.
\textbf{Utility ratio and statistical analysis.} To express downstream performance as a fraction of the achievable ceiling, we report the utility ratio $U_M = M_{\text{synth}} / M_{\text{real}}$, with $M \in \{\text{Dice}, \text{NSD}_{2\mathrm{mm}}\}$, computed pairwise per subject and per class on those subjects whose real-MR score exceeded 0.1 so that the ratio remains meaningful. The utility ratio answers the operational question of how much of the real-MR segmentation utility is preserved by each synthesis. Pairwise comparison against real T2w / FLAIR used the Wilcoxon signed-rank test, and the relation between the four fidelity metrics of Section~2.6 and downstream utility was quantified by Pearson linear and Spearman rank correlation across the 48 experiments.
\subsection{Subgroup analysis: histological grade and reoperation}
We complemented the cohort-level evaluation with a planned subgroup analysis along two clinically meaningful axes: (i) histological grade, low-grade glioma (LGG) versus high-grade glioma (HGG), and (ii) reoperation status (yes / no).
For the synthesis benchmark of Section~2.6, the four image-quality metrics (SSIM, PSNR, MAE, LPIPS) were recomputed within each stratum, and between-stratum differences were tested with the Mann--Whitney $U$ test on per-subject means at $\alpha = 0.05$; multiple-comparison correction was not applied because the four metrics are not independent and the analysis is intentionally exploratory. For the downstream evaluation of Section~2.7, the same per-subject utility ratios $U_M = M_{\text{synth}} / M_{\text{real}}$ ($M \in \{\text{Dice}, \text{NSD}_{2\mathrm{mm}}\}$) of Section~\ref{sec:downstream} were recomputed within each subgroup, with the inclusion criterion (real-T2w score $> 0.1$) applied per stratum. Subgroup sample sizes are summarised in Table~\ref{tab:cohort}; per-subgroup means are reported in Table~\ref{tab:subgroup_synth} (synthesis) and Table~\ref{tab:subgroup_utility} (downstream utility), and visualised in Figures~\ref{fig:subgroup_synth} (synthesis) and~\ref{fig:subgroup_downstream} (downstream).
\section{Results}
\label{sec:results}
\begin{figure}[!htbp]
\centering
\includegraphics[width=\textwidth]{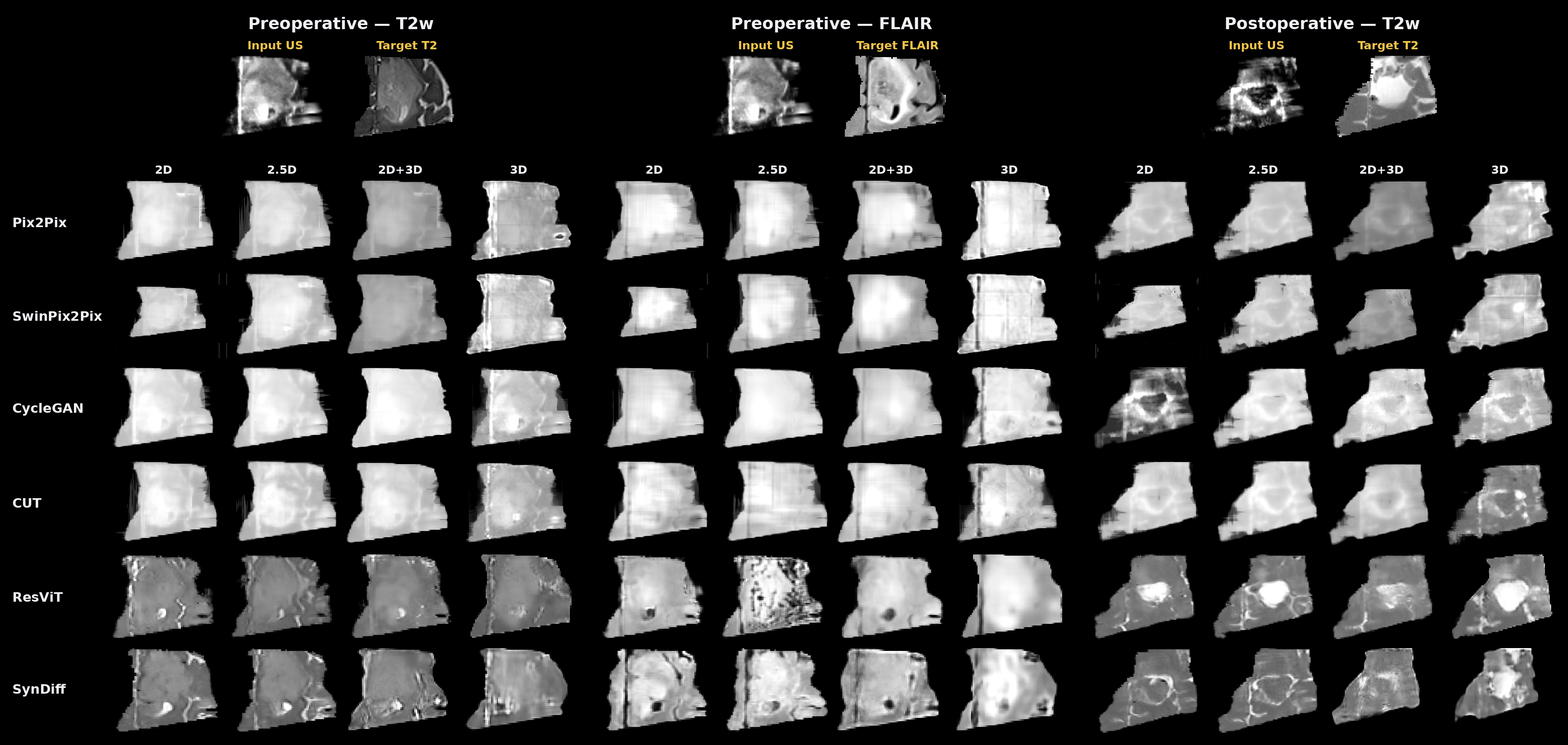}
\caption{Qualitative example of ioUS$\to$MRI synthesis on a representative held-out test case (ReMIND-091; WHO grade~3 astrocytoma, IDH-mutant, no reoperation). Three acquisition settings are shown side by side: preoperative T2w (left), preoperative FLAIR (centre) and postoperative T2w (right). For each setting, the top row gives the input intraoperative ultrasound (Input US) and the acquired reference (Target). Below, each row corresponds to one generator (Pix2Pix, SwinPix2Pix, CycleGAN, CUT, ResViT, SynDiff) and each column to one inference regime (2D, 2.5D, 2D\,+\,3D-refine, full-3D); the preoperative FLAIR block shows the FLAIR channel of the corresponding multi-task (T2w\,+\,FLAIR) run. All panels are displayed on the FOV-cropped grid after intensity normalisation. The GAN family reconstructs smooth, globally plausible tissue but tends to render the postoperative resection cavity as homogeneous parenchyma, whereas ResViT and SynDiff reproduce more of the cavity texture and lesion contrast, consistent with their higher perceptual and downstream-utility scores (Tables~\ref{tab:t2w_single}--\ref{tab:flair_multi}, \ref{tab:utility}).}
\label{fig:qualitative}
\end{figure}

\begin{figure}[!htbp]
\centering
\includegraphics[width=\textwidth]{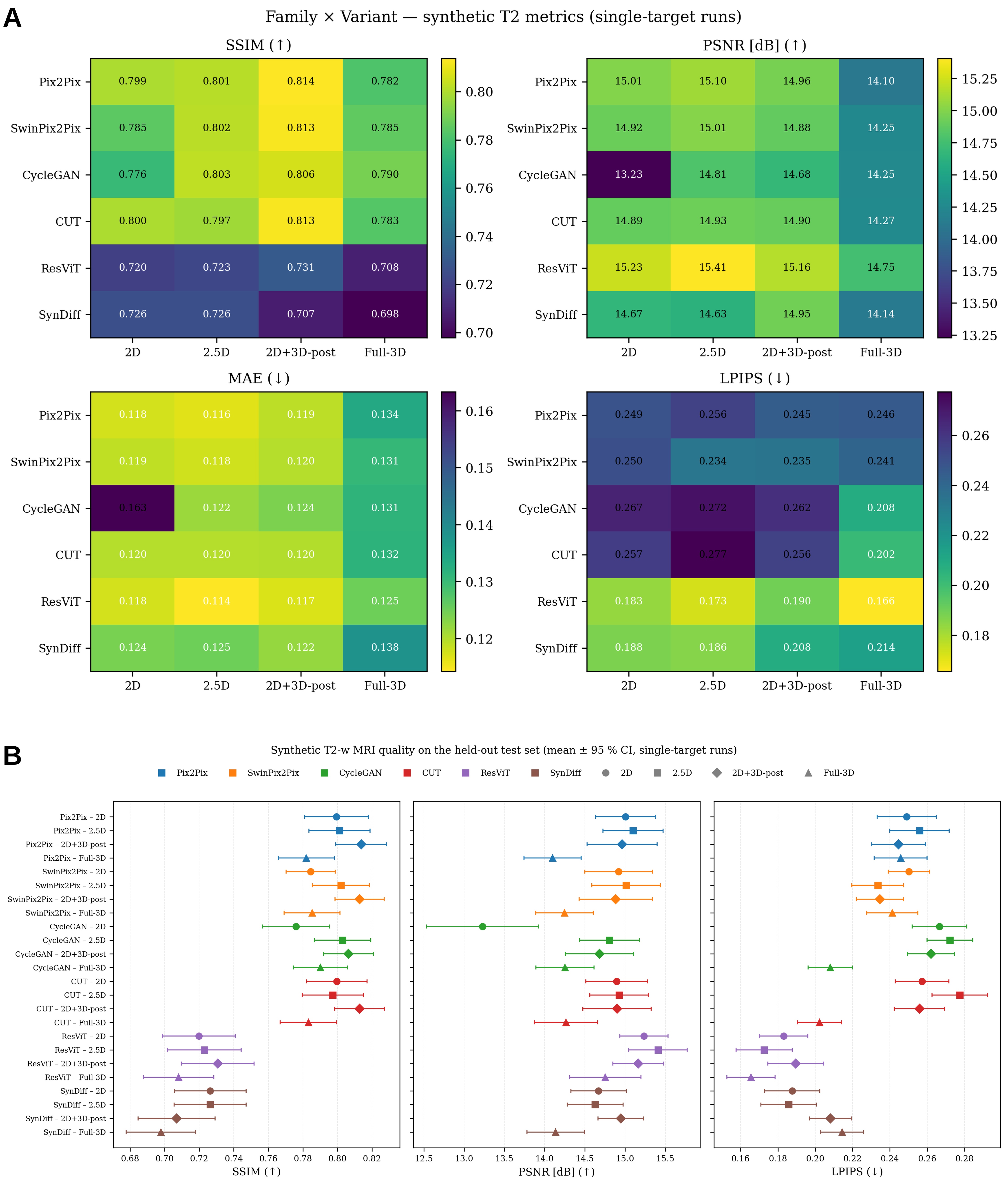}
\caption{Synthetic T2w quality on the held-out test set (single-target runs, $n = 31$ paired studies). \textbf{(A)} Family $\times$ inference-regime heat-map of the mean SSIM, PSNR, MAE and LPIPS; warm colours = better SSIM/PSNR, cool colours = better MAE/LPIPS. \textbf{(B)} Forest plot of the mean $\pm$ 95\,\% Student-$t$ CI; colours encode the architectural family, markers encode the inference regime. The 2D + 3D-refinement variants of the GAN family dominate SSIM, the ResViT family dominates PSNR, and the ResViT family, with SynDiff close behind, dominates LPIPS.}
\label{fig:t2_overview}
\end{figure}

\begin{figure}[!htbp]
\centering
\includegraphics[width=\textwidth]{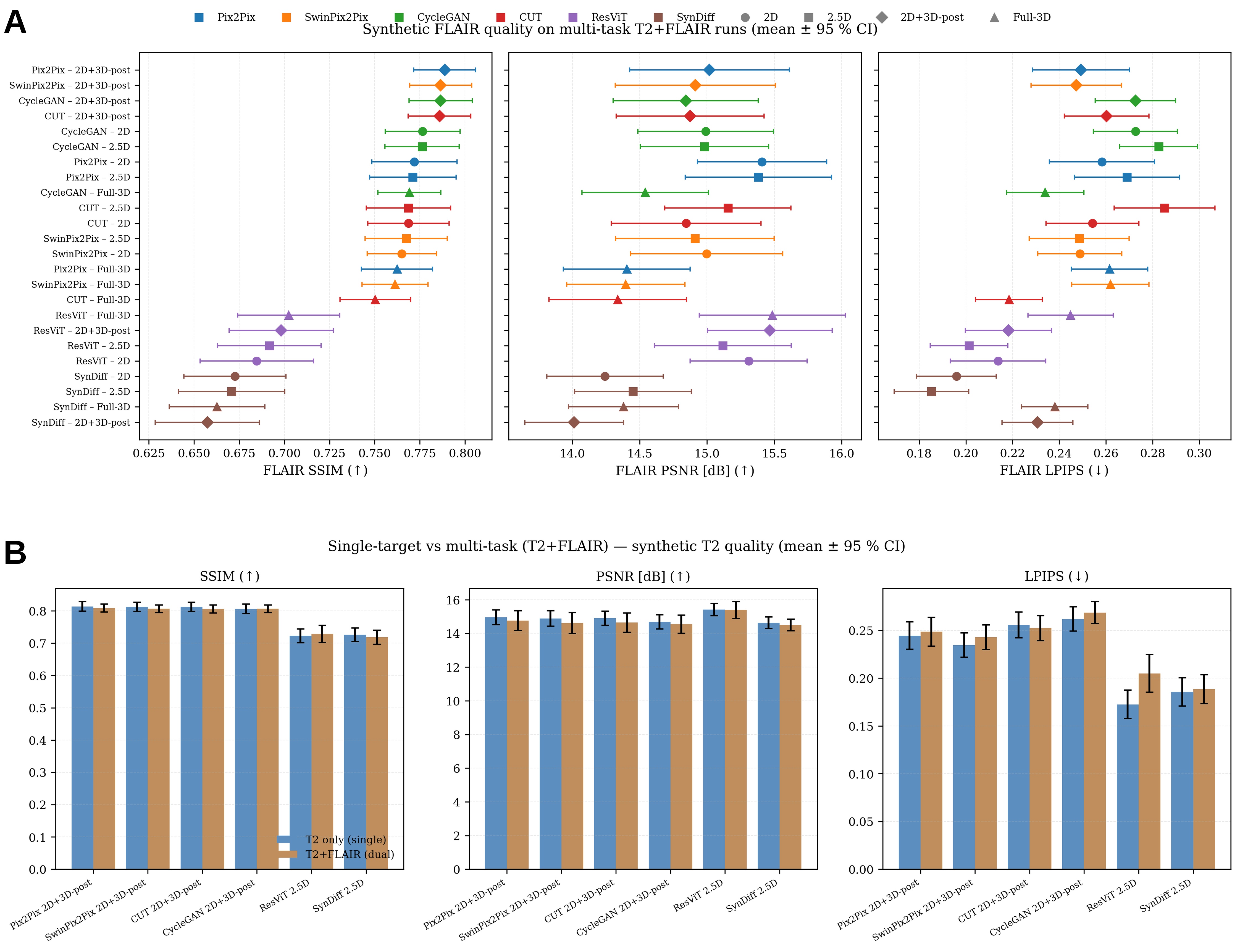}
\caption{Multi-task synthesis quality on the held-out subset of $N = 20$ paired studies containing both T2w and FLAIR. \textbf{(A)} Forest plot of synthetic FLAIR quality (mean $\pm$ 95\,\% Student-$t$ CI). Family colour and regime marker follow Fig.~\ref{fig:t2_overview}B. \textbf{(B)} Single-target (T2w only) versus multi-task (T2w + FLAIR) synthetic T2w quality (mean $\pm$ 95\,\% Student-$t$ CI). Adding FLAIR as a second output channel costs at most 0.010 SSIM and 0.20\,dB PSNR for the GAN family; ResViT loses a similar margin on SSIM but gains marginally on PSNR; SynDiff is the only family for which multi-task supervision does not measurably degrade the T2w metrics.}
\label{fig:multitask}
\end{figure}

Forty-eight experiments were evaluated on the held-out test set: 32 GAN-baseline runs (4 architectures $\times$ 4 inference regimes $\times$ 2 target configurations), 8 ResViT runs (4 regimes $\times$ 2 targets) and 8 SynDiff runs (4 regimes $\times$ 2 targets). Single-target T2w experiments were scored on the full set of $n = 31$ paired studies; multi-task experiments were scored on the subset of $n = 20$ paired studies that contained both T2w and FLAIR. Tables~\ref{tab:t2w_single}--\ref{tab:flair_multi} give the per-experiment mean and 95\,\% Student-$t$ CI for SSIM, PSNR, MAE and LPIPS; Tables~\ref{tab:top_ssim}--\ref{tab:top_lpips} give the cross-family top-10 leaderboards; Table~\ref{tab:pre_post} gives the pre-resection versus post-resection split for the eight strongest models. Figure~\ref{fig:qualitative} shows a representative qualitative example across all experiments, and Figures~\ref{fig:t2_overview} and~\ref{fig:multitask} visualise the rankings.
\begingroup\footnotesize\setlength{\tabcolsep}{3pt}
\begin{longtable}{@{}llrrrr@{}}
\caption{Synthetic T2w quality on the held-out test set, single-target runs. Mean (95\,\% Student-$t$ confidence interval). $n = 31$ paired studies (16 patients).}
\label{tab:t2w_single}\\
\toprule
\textbf{Family} & \textbf{Regime} & \textbf{SSIM $\uparrow$} & \textbf{PSNR $\uparrow$ (dB)} & \textbf{MAE $\downarrow$} & \textbf{LPIPS $\downarrow$} \\
\midrule
\endfirsthead
\multicolumn{6}{l}{\textit{Table~\ref{tab:t2w_single} continued.}}\\
\toprule
\textbf{Family} & \textbf{Regime} & \textbf{SSIM $\uparrow$} & \textbf{PSNR $\uparrow$ (dB)} & \textbf{MAE $\downarrow$} & \textbf{LPIPS $\downarrow$} \\
\midrule
\endhead
\bottomrule
\endfoot
Pix2Pix & 2D & 0.799 (0.781, 0.818) & 15.01 (14.63, 15.38) & 0.118 (0.110, 0.125) & 0.249 (0.233, 0.265) \\
Pix2Pix & 2.5D & 0.801 (0.783, 0.819) & 15.10 (14.72, 15.47) & 0.116 (0.109, 0.124) & 0.256 (0.240, 0.272) \\
Pix2Pix & 2D + 3D-refine & 0.814 (0.799, 0.829) & 14.96 (14.53, 15.40) & 0.119 (0.111, 0.128) & 0.245 (0.230, 0.259) \\
Pix2Pix & Full-3D & 0.782 (0.766, 0.798) & 14.10 (13.74, 14.45) & 0.134 (0.126, 0.141) & 0.246 (0.232, 0.260) \\
SwinPix2Pix & 2D & 0.785 (0.770, 0.799) & 14.92 (14.50, 15.34) & 0.119 (0.111, 0.127) & 0.250 (0.239, 0.261) \\
SwinPix2Pix & 2.5D & 0.802 (0.786, 0.818) & 15.01 (14.58, 15.44) & 0.118 (0.109, 0.126) & 0.234 (0.220, 0.247) \\
SwinPix2Pix & 2D + 3D-refine & 0.813 (0.799, 0.827) & 14.88 (14.43, 15.34) & 0.120 (0.111, 0.128) & 0.235 (0.222, 0.247) \\
SwinPix2Pix & Full-3D & 0.785 (0.769, 0.801) & 14.25 (13.89, 14.60) & 0.131 (0.123, 0.139) & 0.241 (0.228, 0.255) \\
CycleGAN & 2D & 0.776 (0.757, 0.795) & 13.23 (12.53, 13.92) & 0.163 (0.146, 0.181) & 0.267 (0.252, 0.281) \\
CycleGAN & 2.5D & 0.803 (0.787, 0.819) & 14.81 (14.43, 15.18) & 0.122 (0.113, 0.130) & 0.272 (0.260, 0.284) \\
CycleGAN & 2D + 3D-refine & 0.806 (0.792, 0.821) & 14.68 (14.26, 15.10) & 0.124 (0.115, 0.133) & 0.262 (0.249, 0.275) \\
CycleGAN & Full-3D & 0.790 (0.774, 0.806) & 14.25 (13.89, 14.61) & 0.131 (0.123, 0.140) & 0.208 (0.196, 0.220) \\
CUT & 2D & 0.800 (0.782, 0.817) & 14.89 (14.51, 15.28) & 0.120 (0.112, 0.128) & 0.257 (0.243, 0.272) \\
CUT & 2.5D & 0.797 (0.780, 0.815) & 14.93 (14.56, 15.29) & 0.120 (0.112, 0.127) & 0.277 (0.263, 0.292) \\
CUT & 2D + 3D-refine & 0.813 (0.798, 0.827) & 14.90 (14.48, 15.32) & 0.120 (0.111, 0.129) & 0.256 (0.242, 0.269) \\
CUT & Full-3D & 0.783 (0.767, 0.800) & 14.27 (13.87, 14.66) & 0.132 (0.122, 0.141) & 0.202 (0.190, 0.214) \\
ResViT & 2D & 0.720 (0.699, 0.741) & 15.23 (14.93, 15.53) & 0.118 (0.111, 0.124) & 0.183 (0.170, 0.196) \\
ResViT & 2.5D & 0.723 (0.702, 0.744) & 15.41 (15.05, 15.77) & 0.114 (0.107, 0.122) & 0.173 (0.158, 0.188) \\
ResViT & 2D + 3D-refine & 0.731 (0.710, 0.752) & 15.16 (14.85, 15.48) & 0.117 (0.110, 0.124) & 0.190 (0.175, 0.204) \\
ResViT & Full-3D & 0.708 (0.688, 0.728) & 14.75 (14.31, 15.20) & 0.125 (0.116, 0.134) & 0.166 (0.153, 0.178) \\
SynDiff & 2D & 0.726 (0.706, 0.747) & 14.67 (14.33, 15.01) & 0.124 (0.116, 0.132) & 0.188 (0.173, 0.202) \\
SynDiff & 2.5D & 0.726 (0.705, 0.747) & 14.63 (14.28, 14.97) & 0.125 (0.117, 0.133) & 0.186 (0.171, 0.201) \\
SynDiff & 2D + 3D-refine & 0.707 (0.685, 0.729) & 14.95 (14.66, 15.23) & 0.122 (0.115, 0.128) & 0.208 (0.197, 0.219) \\
SynDiff & Full-3D & 0.698 (0.678, 0.718) & 14.14 (13.78, 14.49) & 0.138 (0.130, 0.147) & 0.214 (0.203, 0.226) \\
\end{longtable}
\endgroup

\begingroup\footnotesize\setlength{\tabcolsep}{3pt}
\begin{longtable}{@{}llrrrr@{}}
\caption{Synthetic T2w quality from the multi-task (T2w + FLAIR) runs. Mean (95\,\% Student-$t$ CI). $N = 20$ paired studies that contain both T2w and FLAIR.}
\label{tab:t2w_multi}\\
\toprule
\textbf{Family} & \textbf{Regime} & \textbf{SSIM $\uparrow$} & \textbf{PSNR $\uparrow$ (dB)} & \textbf{MAE $\downarrow$} & \textbf{LPIPS $\downarrow$} \\
\midrule
\endfirsthead
\multicolumn{6}{l}{\textit{Table~\ref{tab:t2w_multi} continued.}}\\
\toprule
\textbf{Family} & \textbf{Regime} & \textbf{SSIM $\uparrow$} & \textbf{PSNR $\uparrow$ (dB)} & \textbf{MAE $\downarrow$} & \textbf{LPIPS $\downarrow$} \\
\midrule
\endhead
\bottomrule
\endfoot
Pix2Pix & 2D & 0.792 (0.775, 0.809) & 14.95 (14.46, 15.43) & 0.118 (0.108, 0.128) & 0.254 (0.237, 0.271) \\
Pix2Pix & 2.5D & 0.792 (0.775, 0.809) & 15.01 (14.46, 15.56) & 0.118 (0.107, 0.130) & 0.268 (0.250, 0.285) \\
Pix2Pix & 2D + 3D-refine & 0.808 (0.796, 0.821) & 14.76 (14.17, 15.35) & 0.121 (0.110, 0.132) & 0.249 (0.234, 0.264) \\
Pix2Pix & Full-3D & 0.780 (0.765, 0.796) & 14.13 (13.67, 14.59) & 0.132 (0.121, 0.143) & 0.260 (0.247, 0.272) \\
SwinPix2Pix & 2D & 0.787 (0.774, 0.800) & 14.61 (14.05, 15.16) & 0.122 (0.110, 0.134) & 0.245 (0.232, 0.257) \\
SwinPix2Pix & 2.5D & 0.788 (0.772, 0.805) & 14.51 (13.89, 15.13) & 0.125 (0.112, 0.138) & 0.248 (0.232, 0.263) \\
SwinPix2Pix & 2D + 3D-refine & 0.807 (0.795, 0.819) & 14.61 (14.00, 15.23) & 0.122 (0.110, 0.134) & 0.243 (0.230, 0.256) \\
SwinPix2Pix & Full-3D & 0.783 (0.770, 0.796) & 14.06 (13.62, 14.51) & 0.131 (0.120, 0.142) & 0.254 (0.241, 0.267) \\
CycleGAN & 2D & 0.797 (0.783, 0.811) & 14.51 (14.02, 15.01) & 0.125 (0.114, 0.136) & 0.269 (0.257, 0.281) \\
CycleGAN & 2.5D & 0.797 (0.782, 0.811) & 14.55 (14.05, 15.05) & 0.124 (0.113, 0.135) & 0.277 (0.265, 0.288) \\
CycleGAN & 2D + 3D-refine & 0.807 (0.795, 0.819) & 14.55 (14.01, 15.08) & 0.124 (0.112, 0.136) & 0.269 (0.257, 0.280) \\
CycleGAN & Full-3D & 0.788 (0.774, 0.801) & 14.25 (13.77, 14.73) & 0.130 (0.119, 0.142) & 0.230 (0.218, 0.242) \\
CUT & 2D & 0.787 (0.770, 0.804) & 14.50 (13.98, 15.01) & 0.127 (0.116, 0.139) & 0.245 (0.230, 0.261) \\
CUT & 2.5D & 0.790 (0.773, 0.806) & 14.80 (14.36, 15.24) & 0.120 (0.110, 0.131) & 0.276 (0.260, 0.293) \\
CUT & 2D + 3D-refine & 0.806 (0.793, 0.818) & 14.64 (14.07, 15.21) & 0.124 (0.111, 0.136) & 0.252 (0.239, 0.266) \\
CUT & Full-3D & 0.773 (0.757, 0.789) & 14.01 (13.52, 14.51) & 0.135 (0.122, 0.147) & 0.217 (0.203, 0.231) \\
ResViT & 2D & 0.714 (0.685, 0.742) & 15.30 (14.88, 15.73) & 0.114 (0.104, 0.123) & 0.209 (0.190, 0.228) \\
ResViT & 2.5D & 0.729 (0.703, 0.756) & 15.39 (14.89, 15.88) & 0.113 (0.103, 0.123) & 0.205 (0.185, 0.225) \\
ResViT & 2D + 3D-refine & 0.730 (0.703, 0.756) & 15.36 (14.90, 15.82) & 0.113 (0.104, 0.123) & 0.230 (0.208, 0.251) \\
ResViT & Full-3D & 0.701 (0.675, 0.726) & 14.81 (14.26, 15.37) & 0.124 (0.112, 0.136) & 0.167 (0.151, 0.183) \\
SynDiff & 2D & 0.717 (0.695, 0.738) & 14.36 (14.03, 14.70) & 0.129 (0.121, 0.136) & 0.198 (0.183, 0.214) \\
SynDiff & 2.5D & 0.718 (0.696, 0.740) & 14.51 (14.16, 14.86) & 0.127 (0.119, 0.135) & 0.189 (0.173, 0.204) \\
SynDiff & 2D + 3D-refine & 0.694 (0.672, 0.716) & 14.05 (13.74, 14.35) & 0.137 (0.131, 0.144) & 0.230 (0.214, 0.245) \\
SynDiff & Full-3D & 0.709 (0.690, 0.729) & 14.33 (13.93, 14.73) & 0.131 (0.122, 0.140) & 0.222 (0.208, 0.236) \\
\end{longtable}
\endgroup
\begingroup\footnotesize\setlength{\tabcolsep}{3pt}
\begin{longtable}{@{}llrrrr@{}}
\caption{Synthetic FLAIR quality from the multi-task (T2w + FLAIR) runs. Mean (95\,\% Student-$t$ CI). $N = 20$.}
\label{tab:flair_multi}\\
\toprule
\textbf{Family} & \textbf{Regime} & \textbf{SSIM $\uparrow$} & \textbf{PSNR $\uparrow$ (dB)} & \textbf{MAE $\downarrow$} & \textbf{LPIPS $\downarrow$} \\
\midrule
\endfirsthead
\multicolumn{6}{l}{\textit{Table~\ref{tab:flair_multi} continued.}}\\
\toprule
\textbf{Family} & \textbf{Regime} & \textbf{SSIM $\uparrow$} & \textbf{PSNR $\uparrow$ (dB)} & \textbf{MAE $\downarrow$} & \textbf{LPIPS $\downarrow$} \\
\midrule
\endhead
\bottomrule
\endfoot
Pix2Pix & 2D & 0.772 (0.748, 0.796) & 15.41 (14.93, 15.89) & 0.128 (0.119, 0.137) & 0.258 (0.236, 0.281) \\
Pix2Pix & 2.5D & 0.771 (0.747, 0.795) & 15.38 (14.84, 15.92) & 0.129 (0.118, 0.139) & 0.269 (0.246, 0.292) \\
Pix2Pix & 2D + 3D-refine & 0.789 (0.772, 0.806) & 15.02 (14.42, 15.61) & 0.130 (0.120, 0.141) & 0.249 (0.229, 0.270) \\
Pix2Pix & Full-3D & 0.762 (0.743, 0.782) & 14.40 (13.93, 14.87) & 0.141 (0.132, 0.150) & 0.262 (0.245, 0.278) \\
SwinPix2Pix & 2D & 0.765 (0.746, 0.784) & 15.00 (14.43, 15.56) & 0.134 (0.124, 0.145) & 0.249 (0.231, 0.267) \\
SwinPix2Pix & 2.5D & 0.767 (0.745, 0.790) & 14.91 (14.32, 15.50) & 0.136 (0.125, 0.147) & 0.249 (0.227, 0.270) \\
SwinPix2Pix & 2D + 3D-refine & 0.787 (0.769, 0.804) & 14.91 (14.32, 15.51) & 0.133 (0.122, 0.143) & 0.247 (0.228, 0.267) \\
SwinPix2Pix & Full-3D & 0.761 (0.743, 0.779) & 14.40 (13.96, 14.83) & 0.140 (0.133, 0.148) & 0.262 (0.245, 0.278) \\
CycleGAN & 2D & 0.777 (0.756, 0.797) & 14.99 (14.49, 15.49) & 0.135 (0.126, 0.145) & 0.273 (0.255, 0.291) \\
CycleGAN & 2.5D & 0.776 (0.756, 0.797) & 14.98 (14.50, 15.46) & 0.135 (0.126, 0.144) & 0.283 (0.266, 0.299) \\
CycleGAN & 2D + 3D-refine & 0.787 (0.769, 0.804) & 14.84 (14.30, 15.38) & 0.133 (0.124, 0.143) & 0.273 (0.255, 0.290) \\
CycleGAN & Full-3D & 0.769 (0.752, 0.787) & 14.54 (14.07, 15.01) & 0.140 (0.131, 0.148) & 0.234 (0.217, 0.251) \\
CUT & 2D & 0.769 (0.746, 0.791) & 14.84 (14.29, 15.40) & 0.139 (0.128, 0.151) & 0.254 (0.234, 0.274) \\
CUT & 2.5D & 0.769 (0.745, 0.792) & 15.16 (14.69, 15.62) & 0.133 (0.123, 0.142) & 0.285 (0.264, 0.307) \\
CUT & 2D + 3D-refine & 0.786 (0.769, 0.803) & 14.87 (14.33, 15.42) & 0.133 (0.124, 0.143) & 0.260 (0.242, 0.278) \\
CUT & Full-3D & 0.750 (0.731, 0.770) & 14.34 (13.83, 14.85) & 0.144 (0.135, 0.153) & 0.219 (0.204, 0.233) \\
ResViT & 2D & 0.685 (0.653, 0.716) & 15.31 (14.87, 15.74) & 0.129 (0.121, 0.136) & 0.214 (0.193, 0.234) \\
ResViT & 2.5D & 0.692 (0.663, 0.720) & 15.12 (14.61, 15.62) & 0.131 (0.123, 0.139) & 0.201 (0.185, 0.218) \\
ResViT & 2D + 3D-refine & 0.698 (0.669, 0.727) & 15.47 (15.00, 15.93) & 0.126 (0.119, 0.134) & 0.218 (0.200, 0.237) \\
ResViT & Full-3D & 0.702 (0.674, 0.731) & 15.48 (14.94, 16.03) & 0.124 (0.116, 0.132) & 0.245 (0.227, 0.263) \\
SynDiff & 2D & 0.673 (0.644, 0.701) & 14.24 (13.81, 14.67) & 0.145 (0.136, 0.154) & 0.196 (0.179, 0.213) \\
SynDiff & 2.5D & 0.671 (0.641, 0.700) & 14.45 (14.02, 14.88) & 0.142 (0.133, 0.151) & 0.185 (0.169, 0.201) \\
SynDiff & 2D + 3D-refine & 0.657 (0.628, 0.686) & 14.01 (13.65, 14.38) & 0.150 (0.142, 0.158) & 0.231 (0.215, 0.246) \\
SynDiff & Full-3D & 0.663 (0.636, 0.689) & 14.38 (13.97, 14.79) & 0.144 (0.135, 0.152) & 0.238 (0.224, 0.252) \\
\end{longtable}
\endgroup

\subsection{Cross-family leaderboards}
Pooling all 48 experiments (single + multi-task), three regimes emerge from the per-metric leaderboards (Table~\ref{tab:top_ssim}--\ref{tab:top_lpips}): (i) the four 2D + 3D-refinement GAN baselines top the SSIM ranking by a clear margin; (ii) the ResViT family, especially ResViT-2.5D and ResViT-2D + 3D-refine, tops the PSNR ranking; (iii) the ResViT family, led by Full-3D, tops the LPIPS ranking, with the SynDiff variants close behind, producing more perceptually realistic synthetic T2w even though they trail the GAN baselines on intensity-domain metrics.
\begin{table}[ht]
\centering
\caption{Top-10 ranking by mean SSIM (synthetic T2w, all single + multi-task runs). Higher is better.}
\label{tab:top_ssim}
\footnotesize
\setlength{\tabcolsep}{4pt}
\begin{tabular}{@{}clllll@{}}
\toprule
\textbf{Rank} & \textbf{Family} & \textbf{Regime} & \textbf{Target} & \textbf{SSIM (95\,\% CI)} & \textbf{$n$} \\
\midrule
1 & Pix2Pix & 2D + 3D-refine & T2w only & 0.814 (0.799, 0.829) & 31 \\
2 & CUT & 2D + 3D-refine & T2w only & 0.813 (0.798, 0.827) & 31 \\
3 & SwinPix2Pix & 2D + 3D-refine & T2w only & 0.813 (0.799, 0.827) & 31 \\
4 & Pix2Pix & 2D + 3D-refine & T2w + FLAIR & 0.808 (0.796, 0.821) & 20 \\
5 & SwinPix2Pix & 2D + 3D-refine & T2w + FLAIR & 0.807 (0.795, 0.819) & 20 \\
6 & CycleGAN & 2D + 3D-refine & T2w + FLAIR & 0.807 (0.795, 0.819) & 20 \\
7 & CycleGAN & 2D + 3D-refine & T2w only & 0.806 (0.792, 0.821) & 31 \\
8 & CUT & 2D + 3D-refine & T2w + FLAIR & 0.806 (0.793, 0.818) & 20 \\
9 & CycleGAN & 2.5D & T2w only & 0.803 (0.787, 0.819) & 31 \\
10 & SwinPix2Pix & 2.5D & T2w only & 0.802 (0.786, 0.818) & 31 \\
\bottomrule
\end{tabular}
\end{table}
\begin{table}[ht]
\centering
\caption{Top-10 ranking by mean PSNR (synthetic T2w). Higher is better.}
\label{tab:top_psnr}
\footnotesize
\setlength{\tabcolsep}{4pt}
\begin{tabular}{@{}clllll@{}}
\toprule
\textbf{Rank} & \textbf{Family} & \textbf{Regime} & \textbf{Target} & \textbf{PSNR dB (95\,\% CI)} & \textbf{$n$} \\
\midrule
1 & ResViT & 2.5D & T2w only & 15.41 (15.05, 15.77) & 31 \\
2 & ResViT & 2.5D & T2w + FLAIR & 15.39 (14.89, 15.88) & 20 \\
3 & ResViT & 2D + 3D-refine & T2w + FLAIR & 15.36 (14.90, 15.82) & 20 \\
4 & ResViT & 2D & T2w + FLAIR & 15.30 (14.88, 15.73) & 20 \\
5 & ResViT & 2D & T2w only & 15.23 (14.93, 15.53) & 31 \\
6 & ResViT & 2D + 3D-refine & T2w only & 15.16 (14.85, 15.48) & 31 \\
7 & Pix2Pix & 2.5D & T2w only & 15.10 (14.72, 15.47) & 31 \\
8 & SwinPix2Pix & 2.5D & T2w only & 15.01 (14.58, 15.44) & 31 \\
9 & Pix2Pix & 2.5D & T2w + FLAIR & 15.01 (14.46, 15.56) & 20 \\
10 & Pix2Pix & 2D & T2w only & 15.01 (14.63, 15.38) & 31 \\
\bottomrule
\end{tabular}
\end{table}
\begin{table}[ht]
\centering
\caption{Top-10 ranking by mean LPIPS-AlexNet (synthetic T2w). Lower is better.}
\label{tab:top_lpips}
\footnotesize
\setlength{\tabcolsep}{4pt}
\begin{tabular}{@{}clllll@{}}
\toprule
\textbf{Rank} & \textbf{Family} & \textbf{Regime} & \textbf{Target} & \textbf{LPIPS (95\,\% CI)} & \textbf{$n$} \\
\midrule
1 & ResViT & Full-3D & T2w only & 0.166 (0.153, 0.178) & 31 \\
2 & ResViT & Full-3D & T2w + FLAIR & 0.167 (0.151, 0.183) & 20 \\
3 & ResViT & 2.5D & T2w only & 0.173 (0.158, 0.188) & 31 \\
4 & ResViT & 2D & T2w only & 0.183 (0.170, 0.196) & 31 \\
5 & SynDiff & 2.5D & T2w only & 0.186 (0.171, 0.201) & 31 \\
6 & SynDiff & 2D & T2w only & 0.188 (0.173, 0.202) & 31 \\
7 & SynDiff & 2.5D & T2w + FLAIR & 0.189 (0.173, 0.204) & 20 \\
8 & ResViT & 2D + 3D-refine & T2w only & 0.190 (0.175, 0.204) & 31 \\
9 & SynDiff & 2D & T2w + FLAIR & 0.198 (0.183, 0.214) & 20 \\
10 & CUT & Full-3D & T2w only & 0.202 (0.190, 0.214) & 31 \\
\bottomrule
\end{tabular}
\end{table}
\subsubsection{ROI-restricted leaderboards}
The same 48 experiments were re-scored on lesion, tumour and cavity ROIs at 0 and 5\,mm dilations (Section~2.6.1). Restricting the metrics to the lesion region substantially rearranges the leaderboard. On the strict 0\,mm lesion mask, the 2D + 3D-refinement GAN variants retain their SSIM lead (SwinPix2Pix-2D + 3D-refine-T2w+FLAIR 0.757, CycleGAN-2D + 3D-refine-T2w+FLAIR 0.753, ResViT-2D + 3D-refine-T2w+FLAIR 0.750; Table~\ref{tab:roi_top}), but the GAN lead collapses once a 5\,mm margin around the lesion is included: ResViT-2D + 3D-refine-T2w+FLAIR (SSIM 0.667) and ResViT-2.5D-T2w+FLAIR (0.657) overtake every GAN variant on SSIM, and ResViT-2.5D-T2w wins on LPIPS (0.120) ahead of ResViT-3D-T2w (0.123) and ResViT-2D-T2w (0.127), with no GAN variant in the top-5 LPIPS at the 5\,mm margin. PSNR and MAE behave consistently with their global counterparts once restricted to the ROI, again placing the ResViT family at the top once a lesion margin is included.
Splitting by ROI sub-component confirms a known but quantitatively informative fact: SSIM and PSNR are sensitive to the underlying intensity statistics of the structure being evaluated. The tumour ROI, a higher-contrast lesion, yields SSIM up to $\sim 0.80$ and PSNR $\sim 17$--18\,dB on the same top methods at 0\,mm dilation (CycleGAN-2D + 3D-refine-T2w+FLAIR 0.796 / 18.08\,dB; ResViT-2D + 3D-refine-T2w+FLAIR 0.793 / 17.45\,dB), while the cavity ROI yields SSIM $\sim 0.70$ and PSNR $\sim 10$\,dB at 0\,mm dilation because cavity content is fluid-like and frequently mis-rendered by the GAN family as smooth tissue (SwinPix2Pix-2D + 3D-refine-T2w+FLAIR cavity 0\,mm: SSIM 0.702, PSNR 10.36\,dB). On cavity at 5\,mm dilation the ranking again becomes ResViT-dominated (ResViT-2D + 3D-refine-T2w+FLAIR SSIM 0.651, PSNR 13.38\,dB), consistent with the transformer family producing more realistic cavity texture even when its absolute intensity statistics differ from the GAN baselines. The pre-resection versus post-resection split (lesion ROI, 0\,mm) is consistent with the global subgroup pattern of Section~3.2: the 2D + 3D-refinement GAN family leads on pre-resection (SwinPix2Pix-2D + 3D-refine-T2w+FLAIR pre 0.764 vs post 0.745), while ResViT-2D + 3D-refine-T2w+FLAIR is the top method on post-resection (0.743); once a 5\,mm margin is included the post-resection ranking flips, with the ResViT family at the top in both phases (ResViT-2D + 3D-refine-T2w+FLAIR pre 0.667, post 0.669). The synthetic FLAIR channel of the multi-task runs follows the same picture: GANs lead at 0\,mm (CycleGAN-2D + 3D-refine-T2w+FLAIR SSIM 0.727), ResViT leads at 5\,mm (ResViT-2D + 3D-refine-T2w+FLAIR SSIM 0.636), and SynDiff variants dominate the LPIPS ranking at 5\,mm (SynDiff-2.5D-T2w+FLAIR 0.135). Taken together, the ROI-restricted analysis confirms the qualitative split of Section~3.1 (GANs lead SSIM, ResViT leads PSNR / LPIPS) but inverts the SSIM ranking in the most clinically meaningful regime (lesion + 5\,mm margin), where ResViT replaces the GAN family at the top of every metric and the apparent SSIM advantage of the GAN baselines is shown to come predominantly from off-target background content.
\begin{table}[ht]
\centering
\caption{ROI-restricted top-5 ranking by mean SSIM on the lesion mask (tumour $\cup$ cavity) at dilation 0\,mm and 5\,mm; T2w channel; $n = 30$ single-target studies and $n = 20$ dual-target studies with available MRI segmentation. The strict 0\,mm mask keeps the 2D + 3D-refinement GAN variants on top, whereas including a 5\,mm margin around the lesion brings the ResViT family to the top.}
\label{tab:roi_top}
\scriptsize
\setlength{\tabcolsep}{3pt}
\begin{tabular}{@{}cllllrrrr@{}}
\toprule
\textbf{Rank} & \textbf{Dilation} & \textbf{Family} & \textbf{Regime} & \textbf{Target} & \textbf{SSIM $\uparrow$} & \textbf{PSNR $\uparrow$ (dB)} & \textbf{LPIPS $\downarrow$} & \textbf{$n$} \\
\midrule
1 & 0\,mm & SwinPix2Pix & 2D + 3D-refine & T2w + FLAIR & 0.757 & 15.36 & 0.105 & 20 \\
2 & 0\,mm & CycleGAN & 2D + 3D-refine & T2w + FLAIR & 0.753 & 14.88 & 0.108 & 20 \\
3 & 0\,mm & ResViT & 2D + 3D-refine & T2w + FLAIR & 0.750 & 14.72 & 0.107 & 20 \\
4 & 0\,mm & Pix2Pix & 2D + 3D-refine & T2w + FLAIR & 0.748 & 14.47 & 0.112 & 20 \\
5 & 0\,mm & CUT & 2D + 3D-refine & T2w + FLAIR & 0.746 & 14.52 & 0.107 & 20 \\
\midrule
1 & 5\,mm & ResViT & 2D + 3D-refine & T2w + FLAIR & 0.667 & 15.64 & 0.154 & 20 \\
2 & 5\,mm & ResViT & 2.5D & T2w + FLAIR & 0.657 & 15.52 & 0.139 & 20 \\
3 & 5\,mm & SwinPix2Pix & 2D + 3D-refine & T2w + FLAIR & 0.655 & 15.32 & 0.191 & 20 \\
4 & 5\,mm & Pix2Pix & 2D + 3D-refine & T2w + FLAIR & 0.653 & 15.10 & 0.202 & 20 \\
5 & 5\,mm & ResViT & 2D + 3D-refine & T2w only & 0.651 & 15.02 & 0.137 & 30 \\
\bottomrule
\end{tabular}
\end{table}
\subsection{Pre-resection versus post-resection synthesis}
\begin{figure}[!htbp]
\centering
\includegraphics[width=\textwidth]{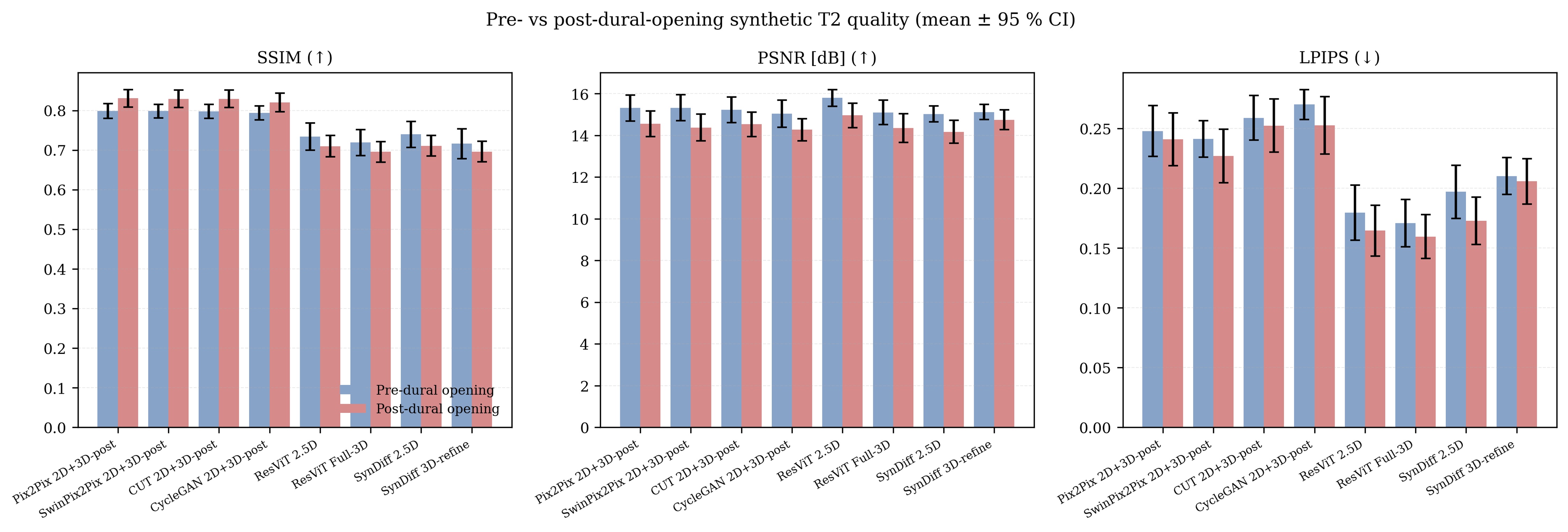}
\caption{Pre-resection versus post-resection synthetic T2w metrics for the eight strongest single-target models on the held-out test set (mean $\pm$ 95\,\% Student-$t$ CI; $n_{\text{pre}} = 16$, $n_{\text{post}} = 15$). Post-resection studies trade SSIM for PSNR in the GAN family; ResViT and SynDiff follow the inverse pattern. LPIPS is consistently lower (better) on post-resection studies for the transformer and diffusion families.}
\label{fig:pre_post}
\end{figure}

We split the test studies by surgical phase and recomputed the metrics for the eight strongest models (Table~\ref{tab:pre_post}, Figure~\ref{fig:pre_post}). The 2D + 3D-refinement GAN family obtains a noticeably higher SSIM on post-resection volumes (e.g.\ Pix2Pix-2D + 3D-refine pre-resection 0.799 [0.780, 0.818] vs.\ post-resection 0.831 [0.809, 0.853]) but loses approximately 0.7--0.9\,dB of PSNR; the transformer and diffusion families show the opposite pattern, with slightly higher SSIM and PSNR on pre-resection studies and a small but consistent advantage on LPIPS for the post-resection ones. Intuitively, the developing resection cavity introduces large low-contrast regions that are trivially easy to match in pixel space (favouring SSIM under the GAN family's aggressive $L_1$ + edge optimisation) but reduce the available high-frequency content that PSNR rewards. The diffusion and transformer families, which optimise a perceptual / probabilistic surrogate, capture the changing texture better and remain ahead on LPIPS in both phases.
\begin{table}[ht]
\centering
\caption{Pre-resection and post-resection synthetic T2w metrics for the eight strongest single-target models. Mean (95\,\% Student-$t$ CI). $n_{\text{pre-resection}} = 16$, $n_{\text{post-resection}} = 15$.}
\label{tab:pre_post}
\resizebox{\textwidth}{!}{%
\begin{tabular}{@{}llrrrrr@{}}
\toprule
\textbf{Model} & \textbf{Phase} & \textbf{SSIM $\uparrow$} & \textbf{PSNR $\uparrow$ (dB)} & \textbf{MAE $\downarrow$} & \textbf{LPIPS $\downarrow$} & \textbf{$n$} \\
\midrule
Pix2Pix 2D + 3D-refine & pre-resection & 0.799 (0.780, 0.818) & 15.31 (14.69, 15.94) & 0.115 (0.104, 0.127) & 0.248 (0.227, 0.269) & 16 \\
Pix2Pix 2D + 3D-refine & post-resection & 0.831 (0.809, 0.853) & 14.56 (13.94, 15.17) & 0.123 (0.109, 0.137) & 0.241 (0.219, 0.263) & 15 \\
SwinPix2Pix 2D + 3D-refine & pre-resection & 0.798 (0.781, 0.816) & 15.32 (14.71, 15.94) & 0.115 (0.103, 0.127) & 0.241 (0.226, 0.257) & 16 \\
SwinPix2Pix 2D + 3D-refine & post-resection & 0.829 (0.807, 0.851) & 14.38 (13.73, 15.02) & 0.125 (0.111, 0.139) & 0.227 (0.205, 0.249) & 15 \\
CycleGAN 2D + 3D-refine & pre-resection & 0.794 (0.776, 0.812) & 15.04 (14.40, 15.69) & 0.120 (0.106, 0.134) & 0.270 (0.258, 0.283) & 16 \\
CycleGAN 2D + 3D-refine & post-resection & 0.820 (0.797, 0.844) & 14.27 (13.74, 14.79) & 0.128 (0.116, 0.141) & 0.253 (0.229, 0.277) & 15 \\
CUT 2D + 3D-refine & pre-resection & 0.798 (0.781, 0.816) & 15.22 (14.60, 15.84) & 0.117 (0.104, 0.129) & 0.259 (0.240, 0.278) & 16 \\
CUT 2D + 3D-refine & post-resection & 0.830 (0.807, 0.852) & 14.53 (13.94, 15.12) & 0.123 (0.110, 0.137) & 0.252 (0.230, 0.275) & 15 \\
ResViT 2.5D & pre-resection & 0.734 (0.700, 0.768) & 15.80 (15.40, 16.20) & 0.108 (0.099, 0.118) & 0.180 (0.157, 0.203) & 16 \\
ResViT 2.5D & post-resection & 0.710 (0.683, 0.737) & 14.96 (14.37, 15.55) & 0.121 (0.108, 0.133) & 0.165 (0.143, 0.186) & 15 \\
ResViT Full-3D & pre-resection & 0.719 (0.686, 0.752) & 15.10 (14.52, 15.68) & 0.119 (0.106, 0.131) & 0.171 (0.151, 0.190) & 16 \\
ResViT Full-3D & post-resection & 0.695 (0.670, 0.721) & 14.36 (13.66, 15.05) & 0.132 (0.118, 0.146) & 0.160 (0.141, 0.178) & 15 \\
SynDiff 2.5D & pre-resection & 0.740 (0.707, 0.773) & 15.02 (14.64, 15.41) & 0.118 (0.106, 0.130) & 0.197 (0.175, 0.219) & 16 \\
SynDiff 2.5D & post-resection & 0.711 (0.685, 0.737) & 14.17 (13.62, 14.72) & 0.132 (0.119, 0.145) & 0.173 (0.153, 0.192) & 15 \\
SynDiff 2D + 3D-refine & pre-resection & 0.716 (0.679, 0.754) & 15.12 (14.76, 15.48) & 0.119 (0.107, 0.131) & 0.210 (0.195, 0.226) & 16 \\
SynDiff 2D + 3D-refine & post-resection & 0.696 (0.671, 0.722) & 14.75 (14.28, 15.22) & 0.124 (0.114, 0.135) & 0.206 (0.187, 0.225) & 15 \\
\bottomrule
\end{tabular}%
}
\end{table}
\subsection{Computational footprint}
Table~\ref{tab:compute} summarises generator parameter counts. Discriminator parameter counts (PatchGAN for the GAN baselines and ResViT; time-conditioned BigGAN-style for SynDiff) are listed because they affect training cost, though not inference. PyTorch latencies were measured using ten warm-up iterations and thirty timed iterations with CUDA events; GAN-family latencies were reported as the axial-only single-pass figure. The full-3D variants are reported as (patch count) $\times$ (per-patch latency) for a representative $\approx 128^3$ assembled volume (eight $96^3$ patches with $24^3$ overlap for ResViT; twenty-seven $64^3$ patches with $16^3$ overlap for SynDiff and the GAN baselines).
\begingroup\footnotesize\setlength{\tabcolsep}{3pt}
\begin{longtable}{@{}llllp{5.0cm}@{}}
\caption{Computational footprint per model: generator parameters ($G$), discriminator parameters ($D$), per-volume axial-only inference latency for an axial volume of $\approx 50$ foreground slices (or sliding-window count $\times$ per-patch latency for the full-3D variants), and peak inference VRAM.}
\label{tab:compute}\\
\toprule
\textbf{Family} & \textbf{Regime} & \textbf{$G$ (M)} & \textbf{$D$ (M)} & \textbf{Latency (s / volume)} \\
\midrule
\endfirsthead
\multicolumn{5}{l}{\textit{Table~\ref{tab:compute} continued.}}\\
\toprule
\textbf{Family} & \textbf{Regime} & \textbf{$G$ (M)} & \textbf{$D$ (M)} & \textbf{Latency (s / volume)} \\
\midrule
\endhead
\bottomrule
\endfoot
Pix2Pix & 2D & 37.0 & 5.5 & $\sim$8--9\,s / volume \\
Pix2Pix & 2.5D & 37.0 & 5.5 & $\sim$8--9\,s / volume \\
Pix2Pix & 2D + 3D-refine & 37.0 + 0.16 & 5.5 & $\sim$8--9\,s + 3D-refiner pass \\
Pix2Pix & Full-3D & 13.97 & 11.05 & sliding-window 27 patches \\
SwinPix2Pix & 2D & 1.44 & 5.5 & $\sim$10\,s / volume \\
SwinPix2Pix & 2.5D & 1.44 & 5.5 & $\sim$10\,s / volume \\
SwinPix2Pix & 2D + 3D-refine & 1.44 + 0.16 & 5.5 & $\sim$10\,s + 3D-refiner pass \\
SwinPix2Pix & Full-3D & 13.97 & 11.05 & sliding-window 27 patches \\
CycleGAN & 2D & 11.37 (inf.) & 2.64 & $\sim$5--6\,s / volume \\
CycleGAN & 2.5D & 11.38 (inf.) & 2.64 & $\sim$5--6\,s / volume \\
CycleGAN & 2D + 3D-refine & 11.37 + 0.16 & 2.64 & $\sim$5--6\,s + 3D-refiner pass \\
CycleGAN & Full-3D & 23.63 (inf.) & 5.29 & sliding-window 27 patches \\
CUT & 2D & 37.09 & 1.32 & $\sim$8--9\,s / volume \\
CUT & 2.5D & 37.10 & 1.32 & $\sim$8--9\,s / volume \\
CUT & 2D + 3D-refine & 37.09 + 0.16 & 1.32 & $\sim$8--9\,s + 3D-refiner pass \\
CUT & Full-3D & --- ($\approx$ Pix2Pix-3D) & 2.64 & sliding-window 27 patches \\
ResViT & 2D & 18.25 & 2.76 & $\approx 0.70$\,s \\
ResViT & 2.5D & 18.25 & 2.77 & $\approx 0.71$\,s \\
ResViT & 2D + 3D-refine & 18.25 + 0.057 & 2.76 & $\approx 0.72$\,s \\
ResViT & Full-3D & 23.78 & 6.22 & $8 \times 96^3$ patches $\approx 0.39$\,s \\
SynDiff & 2D & 12.09 & 7.22 & $\approx 2.15$\,s (4-step sampling) \\
SynDiff & 2.5D & 12.09 & 7.22 & $\approx 2.16$\,s \\
SynDiff & 2D + 3D-refine & 12.09 + 0.21 & 7.22 + 1.56 & $\approx 2.31$\,s (4-step + refiner) \\
SynDiff & Full-3D & 5.42 & 9.12 & $27 \times 64^3$ patches $\approx 4.89$\,s \\
\end{longtable}
\endgroup
\textbf{Training wall-clock.} For the most expensive cells of the matrix: SynDiff 2D / 2.5D took 11--13\,h per 200-epoch run; ResViT-2.5D $\approx 3.9$\,h (Phase 1 100\,ep $\approx 78$\,min + Phase 2 100\,ep $\approx 2.4$\,h); ResViT 2D + 3D-refine $\approx 5.5$--8\,h (including the 30-epoch 3D-refine pass); ResViT full-3D $\approx 6$--6.5\,h; CycleGAN 2D and the 3D GAN baselines 4.5--6\,h; Pix2Pix and CUT 2D $\approx 35$--46\,min; SwinPix2Pix 2D $\approx 70$\,min; SynDiff full-3D 4.7--5.4\,h. The cheap 3D-refinement add-ons ($\approx 10\,000$ steps for the GAN baselines, 30 epochs for ResViT, 100 epochs for SynDiff) cost 15\,min--4\,h on top of the corresponding 2D backbone.
\subsection{Downstream segmentation}
\begin{figure}[!htbp]
\centering
\includegraphics[width=\textwidth]{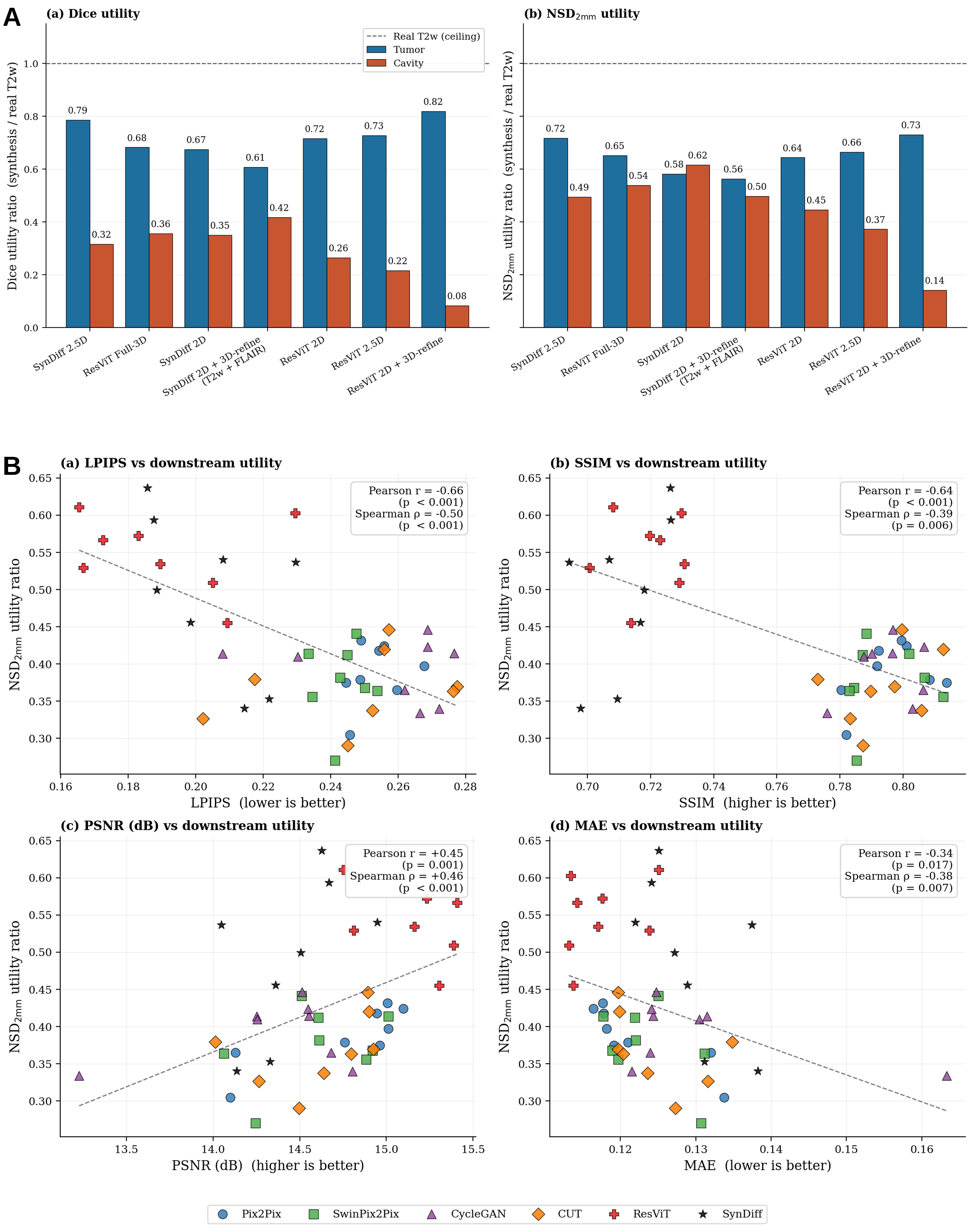}
\caption{Downstream segmentation utility on the held-out test set. \textbf{(A)} Utility ratios (synthesis Dice / real-T2w Dice on the left; synthesis NSD$_{2\mathrm{mm}}$ / real-T2w NSD$_{2\mathrm{mm}}$ on the right), per-subject mean, for the seven strongest synthesis configurations; bars by class (tumour: blue; cavity: orange); dashed line at 1.0 marks the real-T2w ceiling. Tumour utility ranges 0.61--0.82 (Dice) and 0.56--0.73 (NSD); cavity utility is consistently lower (0.08--0.42 Dice, 0.14--0.62 NSD). \textbf{(B)} Relation between image-fidelity metrics (LPIPS, SSIM, PSNR, MAE on T2w) and the downstream NSD$_{2\mathrm{mm}}$ utility ratio across the 48 single-target and multi-task experiments. Each marker is one experiment; colour and shape encode the architectural family. LPIPS is the strongest correlate of downstream utility (Pearson $r = -0.66$); SSIM correlates in the opposite direction to expectation ($r = -0.64$), indicating that pixel-domain similarity does not predict the preservation of segmentation-relevant anatomy.}
\label{fig:downstream}
\end{figure}

Downstream utility was quantified as the per-subject ratio of the synthesis score to the real-T2w score, $U_M = M_{\text{synth}} / M_{\text{real}}$, with $M \in \{\text{Dice}, \text{NSD}_{2\mathrm{mm}}\}$, computed pairwise per study and per class on the test cohort. Real-T2w Dice and NSD$_{2\mathrm{mm}}$, the upper bound, were 0.46 and 0.64 for tumour and 0.39 and 0.68 for resection cavity ($n = 29$ studies per class); subject-level absolute values for the top configurations are reported alongside the utility ratios in Table~\ref{tab:utility}.
\textbf{Pooled utility ranking.} Pooling tumour and cavity, SynDiff-2.5D preserved the largest fraction of real-T2w performance ($U_{\text{Dice}} = 0.55$, $U_{\text{NSD}} = 0.61$), closely followed by ResViT full-3D (0.52 / 0.59) and SynDiff 2D (0.51 / 0.60). All seven highest-utility configurations belonged to the ResViT or SynDiff families; no GAN baseline appeared at the top of either ranking.
\textbf{Per-class utility.} Utility ratios were systematically higher for tumour than for resection cavity across every top configuration (Figure~\ref{fig:downstream}A, Table~\ref{tab:utility}). For tumour, ResViT 2D + 3D-refine achieved the highest utility ($U_{\text{Dice}} = 0.82$, $U_{\text{NSD}} = 0.73$), with SynDiff 2.5D and ResViT 2.5D close behind (0.79 / 0.72 and 0.73 / 0.66, respectively). For resection cavity, SynDiff 2D + 3D-refine (T2w + FLAIR) led on Dice utility (0.42) and SynDiff 2D led on NSD$_{2\mathrm{mm}}$ utility (0.62); ResViT full-3D was the most balanced cavity model (0.36 / 0.54). Notably, ResViT 2D + 3D-refine, the strongest model for tumour utility, was simultaneously the weakest for cavity ($U_{\text{Dice}} = 0.08$, $U_{\text{NSD}} = 0.14$), showing that strong tumour synthesis does not imply faithful reproduction of the resection cavity.

\textbf{Lesion class (tumour $\cup$ cavity).} Treating the two foreground labels as a single class places the question of localisation alongside the question of sub-class identification (Section~2.7). On Seg-T2w ($n = 29$ paired studies), the real-T2w lesion ceiling is Dice 0.662 (NSD$_{2\mathrm{mm}}$ 0.72, HD95 median 5.4\,mm), higher than the per-class ceilings of tumour (0.46) and cavity (0.39) alone, confirming that part of the per-class loss reflects tumour$\leftrightarrow$cavity confusion at the shared boundary rather than a localisation failure. The best synthetic configurations recover 61\,\% of the real-T2w lesion Dice: ResViT-2D + 3D-refine-T2w-from-single (Dice 0.407, NSD$_{2\mathrm{mm}}$ 0.40), followed by ResViT-2.5D-T2w-from-single (0.376 / 0.39) and SynDiff-2D-T2w (0.369 / 0.39); no GAN baseline appears in the top-eight lesion ranking. By phase the gap mirrors the per-class pattern: pre-resection lesion Dice from synthesis 0.55 (ResViT-2D + 3D-refine-T2w-from-single) versus 0.74 from real T2w (74\,\% retention); post-resection 0.30 from ResViT-3D-T2w-from-single versus 0.57 from real T2w (52\,\% retention). On Seg-FLAIR ($n = 19$), the lesion ceiling is Dice 0.511 and the best synthetic configurations preserve a larger fraction of this ceiling than on T2w (ResViT-2.5D-FLAIR Dice 0.373, 73\,\% retention; ResViT-3D-FLAIR 0.366; SynDiff-2D-FLAIR 0.356). Two practical implications follow. First, synthetic FLAIR retains downstream utility more consistently than synthetic T2w in this benchmark (73\,\% vs 61\,\% on the lesion class), suggesting that adding FLAIR as a second synthesis target is operationally worthwhile despite the slightly lower per-voxel fidelity reported in Tables~\ref{tab:t2w_multi}--\ref{tab:flair_multi}. Second, the lesion ranking is dominated by ResViT and SynDiff exactly as on the per-class tumour ranking: paired Wilcoxon tests against real-T2w lesion Dice remain significant for every synthetic source ($p \leq 1.1 \times 10^{-5}$), but the magnitude of the gap ($\sim 0.26$ Dice on T2w; $\sim 0.14$ Dice on FLAIR) is smaller than the gap measured on tumour alone, indicating that an appreciable share of the apparent per-class loss is sub-class boundary confusion rather than missed disease.
\textbf{Statistical comparison with real T2w.} Paired Wilcoxon tests against real T2w were significant for both Dice and NSD$_{2\mathrm{mm}}$ in every top configuration ($p \leq 0.013$ for tumour, $p < 0.001$ for cavity). The class asymmetry is largely driven by the post-resection subset, in which even real-T2w segmentation degrades because residual tumour fragments are often sub-cm$^3$ and cavity boundaries are heterogeneous in low-frequency content. The pre-resection subset, in which tumour boundaries on real T2w are unambiguous, is well preserved by all top configurations (tumour absolute Dice 0.36--0.46 from synthesis versus 0.56 from real T2w on the same subjects).
\textbf{Fidelity-to-utility correlation.} Across the 48 single-target and multi-task experiments, perceptual fidelity predicted downstream utility (Figure~\ref{fig:downstream}B). LPIPS-AlexNet was the strongest correlate of the NSD$_{2\mathrm{mm}}$ utility ratio (Pearson $r = -0.66$, $p < 0.001$; Spearman $\rho = -0.50$, $p < 0.001$) and the strongest correlate of the Dice utility ratio ($r = -0.62$, $p < 0.001$): lower (better) LPIPS corresponded to higher (better) downstream utility. PSNR was a weaker but significant positive correlate of NSD$_{2\mathrm{mm}}$ utility ($r = +0.45$, $p = 0.001$) and uncorrelated with Dice utility ($r = +0.21$, $p = 0.15$); MAE was marginal ($r = -0.34$, $p = 0.02$). Counter-intuitively, SSIM correlated negatively with downstream utility ($r = -0.64$ with NSD$_{2\mathrm{mm}}$ utility, $p < 0.001$): the 2D + 3D-refinement GAN variants that lead the SSIM ranking preserved less downstream utility than the ResViT and SynDiff variants that lead on LPIPS. This pattern is consistent with the documented tendency of SSIM to reward over-smoothed reconstructions in which the high-frequency anatomical detail on which downstream segmentation depends is suppressed.

\begingroup\footnotesize\setlength{\tabcolsep}{4pt}
\begin{longtable}{@{}p{4.4cm}lcccccc@{}}
\caption{Downstream segmentation utility by class for the seven strongest synthesis configurations. The utility ratio $U_M = M_{\text{synth}} / M_{\text{real}}$ is the primary endpoint; absolute Dice and NSD$_{2\mathrm{mm}}$ and median HD95 are given alongside for transparency. Per-subject scores were averaged across only those classes for which a ground-truth structure was present.}
\label{tab:utility}\\
\toprule
\textbf{Model} & \textbf{Class} & \textbf{$n$} & \textbf{$U_{\text{Dice}}$} & \textbf{$U_{\text{NSD2mm}}$} & \textbf{Dice (abs)} & \textbf{NSD2mm (abs)} & \textbf{HD95 (mm)} \\
\midrule
\endfirsthead
\multicolumn{8}{l}{\textit{Table~\ref{tab:utility} continued.}}\\
\toprule
\textbf{Model} & \textbf{Class} & \textbf{$n$} & \textbf{$U_{\text{Dice}}$} & \textbf{$U_{\text{NSD2mm}}$} & \textbf{Dice (abs)} & \textbf{NSD2mm (abs)} & \textbf{HD95 (mm)} \\
\midrule
\endhead
\bottomrule
\endfoot
SynDiff 2.5D & tumour & 16 & 0.79 & 0.72 & 0.434 & 0.444 & 7.7 \\
SynDiff 2.5D & cavity & 13 & 0.32 & 0.49 & 0.224 & 0.361 & 11.5 \\
ResViT Full-3D & tumour & 16 & 0.68 & 0.65 & 0.379 & 0.412 & 6.7 \\
ResViT Full-3D & cavity & 13 & 0.36 & 0.54 & 0.268 & 0.424 & 11.3 \\
SynDiff 2D & tumour & 16 & 0.67 & 0.58 & 0.395 & 0.385 & 9.3 \\
SynDiff 2D & cavity & 13 & 0.35 & 0.62 & 0.243 & 0.449 & 9.9 \\
SynDiff 2D + 3D-refine (T2w+FLAIR) & tumour & 16 & 0.61 & 0.56 & 0.361 & 0.375 & 10.7 \\
SynDiff 2D + 3D-refine (T2w+FLAIR) & cavity & 13 & 0.42 & 0.50 & 0.281 & 0.354 & 13.3 \\
ResViT 2D & tumour & 16 & 0.72 & 0.64 & 0.407 & 0.414 & 10.8 \\
ResViT 2D & cavity & 13 & 0.26 & 0.45 & 0.196 & 0.365 & 11.6 \\
ResViT 2.5D & tumour & 16 & 0.73 & 0.66 & 0.425 & 0.429 & 8.0 \\
ResViT 2.5D & cavity & 13 & 0.22 & 0.37 & 0.167 & 0.298 & 15.7 \\
ResViT 2D + 3D-refine & tumour & 16 & 0.82 & 0.73 & 0.463 & 0.465 & 10.3 \\
ResViT 2D + 3D-refine & cavity & 13 & 0.08 & 0.14 & 0.061 & 0.111 & 17.0 \\
\end{longtable}
\endgroup

\subsection{Subgroup analysis: histological grade and reoperation}
\begin{figure}[!htbp]
\centering
\includegraphics[width=\textwidth]{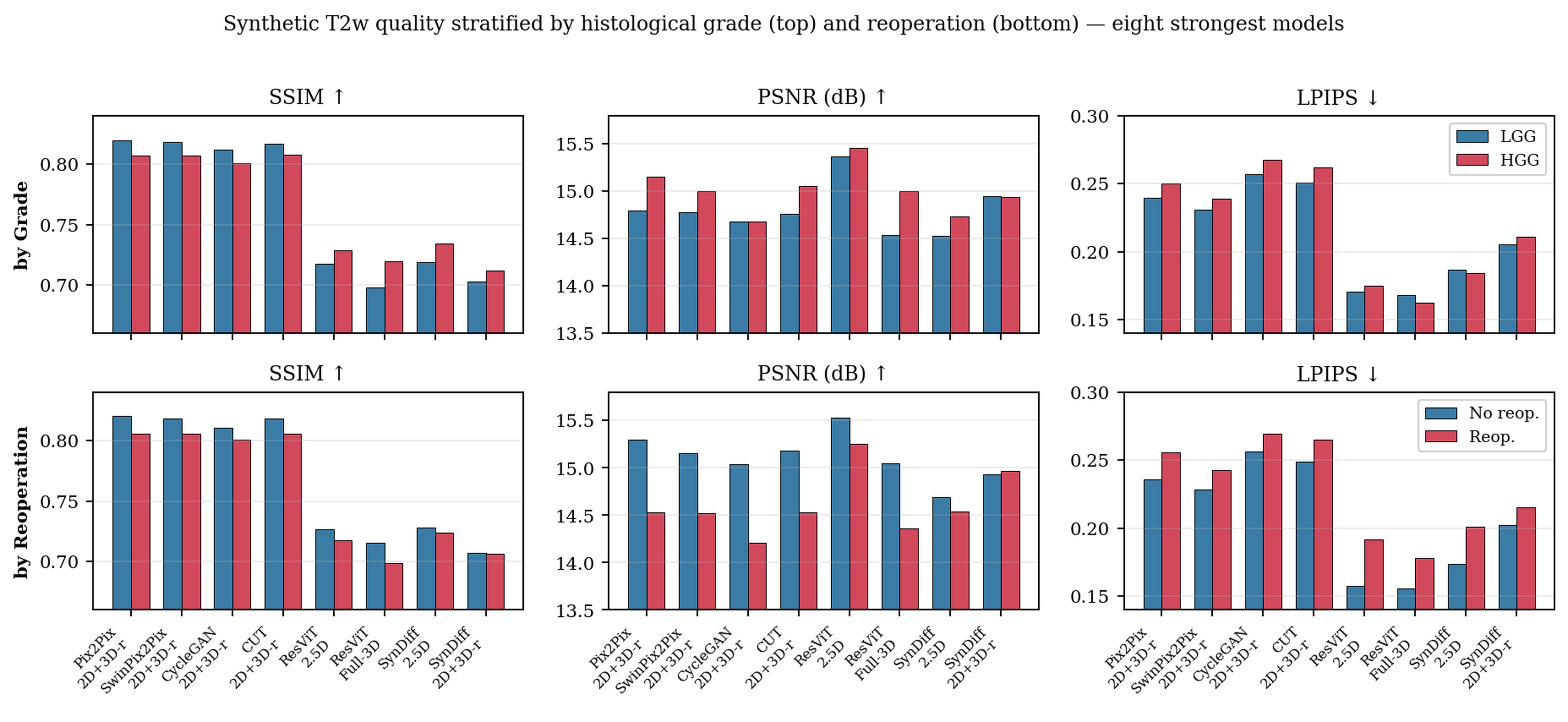}
\caption{Synthetic T2w quality stratified by histological grade (top row) and by reoperation history (bottom row) for the eight strongest models. Bars are the per-subgroup mean over per-subject means. Grade is metric-neutral; reoperation reduces PSNR for the GAN family and ResViT-full-3D, and LPIPS for the ResViT family, with several individual significant effects (Mann--Whitney $U$, $p < 0.05$; see main text).}
\label{fig:subgroup_synth}
\end{figure}

\begin{figure}[!htbp]
\centering
\includegraphics[width=\textwidth]{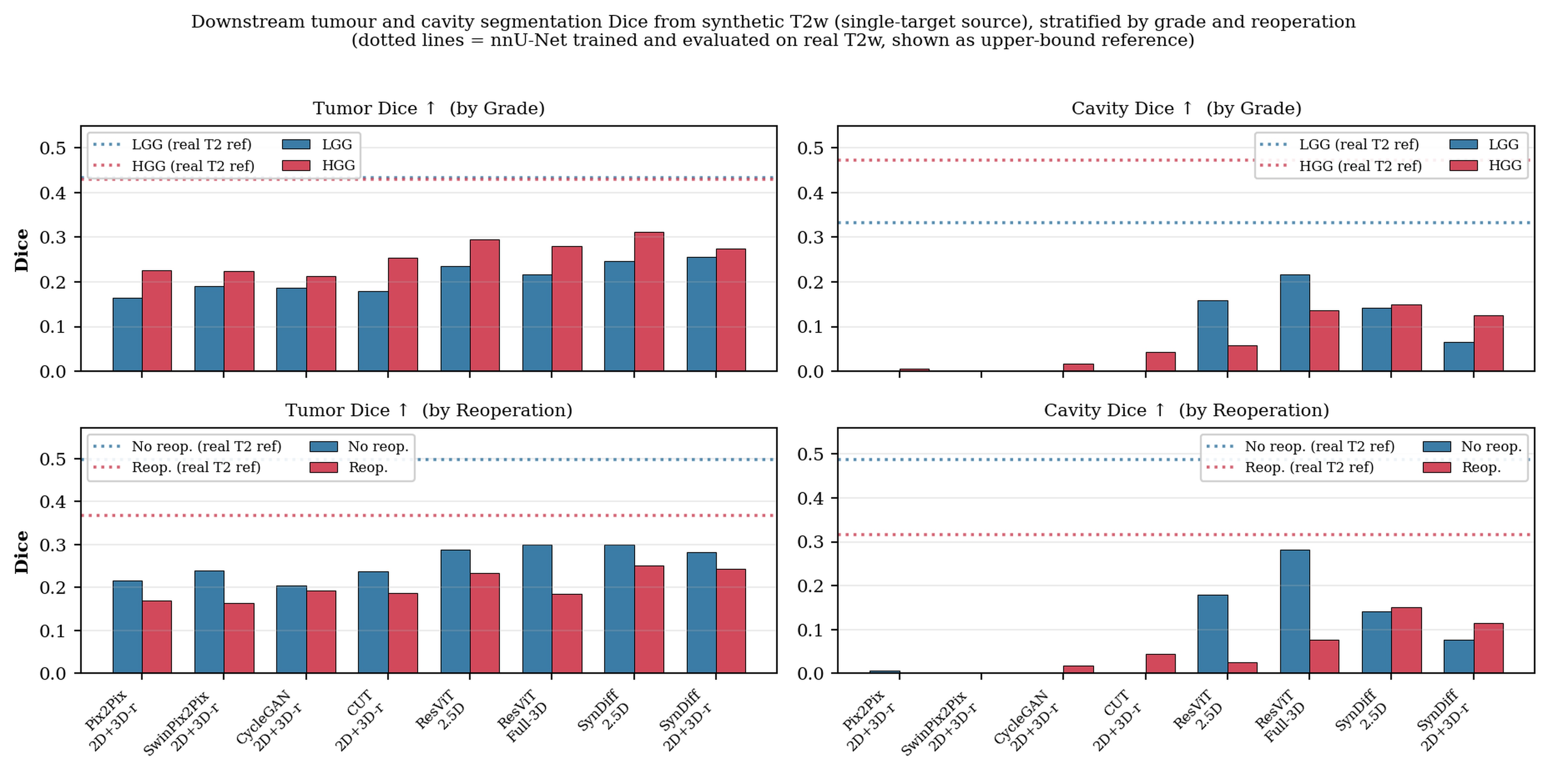}
\caption{Downstream tumour and cavity segmentation Dice (single-target source) for the seven strongest models, stratified by histological grade (top) and by reoperation history (bottom). Bars are the per-subgroup mean over per-study Dice. Dotted horizontal lines mark the real-T2w upper bound (nnU-Net trained and evaluated on the real T2w of the same subjects in the same stratum). For most synthesis sources, tumour Dice is higher in the LGG stratum than in the HGG stratum, paralleling the utility ratios reported in Table~\ref{tab:subgroup_utility} (with SynDiff-2.5D and ResViT-full-3D essentially grade-neutral). Cavity Dice is near zero for the GAN family because the developing resection cavity is reconstructed as smooth tissue rather than as a fluid-like void, while the LPIPS-leading models (ResViT-full-3D, SynDiff-2.5D, SynDiff-2D+3D-refine) retain measurable cavity structure. Reoperation reduces tumour Dice uniformly; for cavity, the effect is mixed---ResViT models degrade further while SynDiff models recover better cavity Dice in reoperation cases.}
\label{fig:subgroup_downstream}
\end{figure}

The 16 test patients of ReMIND comprise 9 LGG and 7 HGG, and 9 first-time interventions versus 7 reoperations (Table~\ref{tab:cohort}). The four resulting joint cells are populated by 5, 4, 4 and 3 patients, respectively, so subgroup confidence intervals are wide; we therefore emphasise the cross-method ranking and the direction of the effect rather than absolute subgroup means.
\begingroup\footnotesize\setlength{\tabcolsep}{3pt}
\begin{longtable}{@{}p{2.0cm}p{3.0cm}c c p{4.0cm} c c@{}}
\caption{Composition of the held-out test cohort.}
\label{tab:cohort}\\
\toprule
\textbf{ID} & \textbf{Histopathology} & \textbf{WHO} & \textbf{IDH} & \textbf{Location} & \textbf{Reoperation} & \textbf{Grade} \\
\midrule
\endfirsthead
\multicolumn{7}{l}{\textit{Table~\ref{tab:cohort} continued.}}\\
\toprule
\textbf{ID} & \textbf{Histopathology} & \textbf{WHO} & \textbf{IDH} & \textbf{Location} & \textbf{Reoperation} & \textbf{Grade} \\
\midrule
\endhead
\bottomrule
\endfoot
ReMIND-003 & Oligodendroglioma & 2 & Mut & Left posterior temporal & No & Low \\
ReMIND-004 & DNET & 1 & WT & Left inferior parietal & No & Low \\
ReMIND-023 & Oligodendroglioma & 2 & Mut & Right frontal & Yes & Low \\
ReMIND-031 & Glioblastoma & 4 & WT & Right skull-base temporal & No & High \\
ReMIND-032 & Glioblastoma & 4 & WT & Right frontal caudate & No & High \\
ReMIND-034 & Astrocytoma & 4 & Mut & Right temporal & Yes & High \\
ReMIND-045 & Glioblastoma & 4 & WT & Right temporal & No & High \\
ReMIND-056 & Oligodendroglioma & 2 & Mut & Left temporal & No & Low \\
ReMIND-077 & Astrocytoma & 3 & Mut & Right parietal & Yes & High \\
ReMIND-079 & Astrocytoma & 3 & Mut & Left frontal & Yes & High \\
ReMIND-087 & Oligodendroglioma & 2 & Mut & Left frontal & Yes & Low \\
ReMIND-091 & Astrocytoma & 3 & Mut & Right frontal & No & High \\
ReMIND-096 & Astrocytoma & 2 & Mut & Left parietal & No & Low \\
ReMIND-102 & Low-grade glioma & n/a & WT & Left parieto-occipital & Yes & Low \\
ReMIND-107 & Oligodendroglioma & 2 & Mut & Left frontal (septal) & Yes & Low \\
ReMIND-109 & Oligodendroglioma & 2 & Mut & Left middle temporal gyrus & No & Low \\
\end{longtable}
\endgroup
\textbf{Synthesis quality by subgroup.} No Mann--Whitney $U$ comparison between LGG and HGG reached $p < 0.05$ on any synthesis metric for any of the eight strongest models (Table~\ref{tab:subgroup_synth}, Figure~\ref{fig:subgroup_synth}, top row). LGG and HGG means differ by less than 0.015 SSIM, less than 0.5\,dB PSNR and less than 0.01 LPIPS for every model. Synthesis quality is therefore essentially independent of histological grade, an encouraging property for clinical generalisation. The reoperation stratum, by contrast, degraded synthesis on the perceptual and intensity-fidelity axes: PSNR was 0.6--0.8\,dB lower for reoperation cases in the GAN family and ResViT-Full-3D (ResViT-2.5D and the SynDiff variants were largely insensitive), considering all 48 experiments, statistically significant individual reoperation effects appeared on PSNR for CUT-Full-3D ($p = 0.007$), SwinPix2Pix-Full-3D ($p = 0.011$), CycleGAN-2.5D / Full-3D / 2D + 3D-refine ($p \approx 0.03$--0.04) and Pix2Pix-Full-3D ($p = 0.036$); on MAE for SwinPix2Pix-Full-3D ($p = 0.019$) and Pix2Pix-2D + 3D-refine ($p = 0.045$); and on LPIPS for CycleGAN-Full-3D ($p = 0.04$), ResViT-2D and ResViT-2.5D ($p \approx 0.04$--0.05). SSIM was essentially insensitive to reoperation status---a useful warning that SSIM alone misses subgroup effects that LPIPS and PSNR detect. (Table~\ref{tab:subgroup_synth} shows only the eight strongest models and marks the two of these effects that fall within that subset).
\begin{table}[ht]
\centering
\caption{Synthetic T2w quality of the eight strongest models, stratified by histological grade (LGG vs HGG; $n_{\text{LGG}} = 16$, $n_{\text{HGG}} = 15$ paired studies) and by reoperation history (No vs Yes; $n_{\text{No}} = 18$, $n_{\text{Yes}} = 13$). Mean SSIM / PSNR / LPIPS over per-subject means. No grade-stratum comparison reached significance (Mann--Whitney $U$, all $p > 0.05$); reoperation-stratum significance is marked with $\dagger$ ($p < 0.05$).}
\label{tab:subgroup_synth}
\resizebox{\textwidth}{!}{%
\begin{tabular}{@{}l*{6}{c}@{}}
\toprule
\textbf{Model} & \textbf{SSIM (L/H)} & \textbf{PSNR (L/H)} & \textbf{LPIPS (L/H)} & \textbf{SSIM (N/Y)} & \textbf{PSNR (N/Y)} & \textbf{LPIPS (N/Y)} \\
\midrule
Pix2Pix 2D + 3D-refine    & 0.820 / 0.807 & 14.80 / 15.15 & 0.240 / 0.250 & 0.820 / 0.806 & 15.29 / 14.53 & 0.236 / 0.256 \\
SwinPix2Pix 2D + 3D-refine & 0.818 / 0.807 & 14.78 / 15.00 & 0.231 / 0.239 & 0.818 / 0.806 & 15.16 / 14.52 & 0.228 / 0.243 \\
CycleGAN 2D + 3D-refine   & 0.812 / 0.800 & 14.68 / 14.68 & 0.257 / 0.268 & 0.810 / 0.801 & 15.04 / 14.21$\dagger$ & 0.256 / 0.269 \\
CUT 2D + 3D-refine        & 0.817 / 0.808 & 14.76 / 15.06 & 0.251 / 0.262 & 0.818 / 0.806 & 15.18 / 14.53 & 0.249 / 0.265 \\
ResViT 2.5D               & 0.718 / 0.729 & 15.36 / 15.46 & 0.171 / 0.175 & 0.727 / 0.718 & 15.53 / 15.25 & 0.158 / 0.192$\dagger$ \\
ResViT Full-3D            & 0.698 / 0.719 & 14.54 / 15.00 & 0.168 / 0.162 & 0.715 / 0.699 & 15.05 / 14.36 & 0.156 / 0.178 \\
SynDiff 2.5D              & 0.719 / 0.734 & 14.53 / 14.73 & 0.187 / 0.184 & 0.728 / 0.724 & 14.69 / 14.54 & 0.174 / 0.201 \\
SynDiff 2D + 3D-refine    & 0.703 / 0.712 & 14.95 / 14.94 & 0.206 / 0.211 & 0.707 / 0.707 & 14.93 / 14.97 & 0.202 / 0.216 \\
\bottomrule
\end{tabular}%
}
\end{table}
\textbf{Downstream segmentation utility by subgroup.} The downstream utility ratios of Section~3.4, recomputed within each subgroup (Table~\ref{tab:subgroup_utility}, Figure~\ref{fig:subgroup_downstream}), reveal a consistent and clinically interpretable pattern. For tumour, the utility ratio is higher in the LGG stratum than in the HGG stratum for most top configurations (range 0.13--0.31 across models), with SynDiff-2.5D and ResViT-full-3D essentially grade-neutral on tumour utility. For resection cavity, the LGG advantage is larger and consistent across every model ($\approx 0.10$--0.30 higher LGG $U_{\text{Dice}}$), consistent with LGG cavities in this cohort being more contrast-distinct in ioUS than the diffuse infiltrative HGG cavities. Reoperation reduces tumour utility uniformly---the drop in tumour $U_{\text{Dice}}$ is 0.15--0.34 across all top configurations (e.g.\ ResViT 2D + 3D-refine: $0.89 \to 0.74$; SynDiff 2.5D: $0.90 \to 0.68$). For cavity, the reoperation effect is mixed: ResViT 2D + 3D-refine and ResViT 2.5D retain near-zero cavity utility regardless of reoperation ($U_{\text{Dice}} = 0.07$ and $0.05$ in the reoperation stratum), whereas the SynDiff family recovers cavity utility better in reoperation cases than in first-time cases than in first-time cases (SynDiff 2D + 3D-refine T2w+FLAIR cavity $0.30 \to 0.60$; SynDiff 2.5D $0.24 \to 0.43$). These patterns reinforce the conclusion of Section~3.4 that the SSIM-leading and the LPIPS-leading configurations have different downstream profiles, and add a clinical layer: the choice of synthesis family should be conditioned not only on the surgical phase (Section~3.2) but also on the patient's grade and surgical history.
\begin{table}[ht]
\centering
\caption{Downstream tumour and cavity utility ratios ($U_{\text{Dice}} = \text{Dice}_{\text{synth}} / \text{Dice}_{\text{real}}$, mean per subgroup) for the seven strongest synthesis configurations of Section~3.4, stratified by histological grade and by reoperation. Subjects are included in each cell only if their real-T2w Dice exceeds 0.1. The real-T2w upper bound for the same subjects, which differs per stratum, is reported as Dice (real) for direct interpretation.}
\label{tab:subgroup_utility}
\resizebox{\textwidth}{!}{%
\begin{tabular}{@{}llccccl@{}}
\toprule
\textbf{Model} & \textbf{Class} & \textbf{$U_{\text{Dice}}$ LGG} & \textbf{$U_{\text{Dice}}$ HGG} & \textbf{$U_{\text{Dice}}$ No-reop} & \textbf{$U_{\text{Dice}}$ Reop} & \textbf{Real-T2w Dice (LGG/HGG; No/Yes)} \\
\midrule
SynDiff 2.5D                       & tumour & 0.80 & 0.77 & 0.90 & 0.68 & 0.43 / 0.43 ; 0.50 / 0.37 \\
SynDiff 2.5D                       & cavity & 0.39 & 0.27 & 0.24 & 0.43 & 0.33 / 0.47 ; 0.49 / 0.32 \\
ResViT Full-3D                     & tumour & 0.66 & 0.70 & 0.83 & 0.54 & 0.43 / 0.43 ; 0.50 / 0.37 \\
ResViT Full-3D                     & cavity & 0.56 & 0.23 & 0.45 & 0.21 & 0.33 / 0.47 ; 0.49 / 0.32 \\
SynDiff 2D                         & tumour & 0.81 & 0.57 & 0.82 & 0.53 & 0.43 / 0.43 ; 0.50 / 0.37 \\
SynDiff 2D                         & cavity & 0.49 & 0.26 & 0.29 & 0.45 & 0.33 / 0.47 ; 0.49 / 0.32 \\
SynDiff 2D + 3D-refine (T2w+FLAIR) & tumour & 0.78 & 0.47 & 0.78 & 0.44 & 0.43 / 0.43 ; 0.50 / 0.37 \\
SynDiff 2D + 3D-refine (T2w+FLAIR) & cavity & 0.58 & 0.31 & 0.30 & 0.60 & 0.33 / 0.47 ; 0.49 / 0.32 \\
ResViT 2D                          & tumour & 0.84 & 0.62 & 0.82 & 0.61 & 0.43 / 0.43 ; 0.50 / 0.37 \\
ResViT 2D                          & cavity & 0.45 & 0.15 & 0.33 & 0.16 & 0.33 / 0.47 ; 0.49 / 0.32 \\
ResViT 2.5D                        & tumour & 0.80 & 0.67 & 0.86 & 0.59 & 0.43 / 0.43 ; 0.50 / 0.37 \\
ResViT 2.5D                        & cavity & 0.40 & 0.10 & 0.32 & 0.05 & 0.33 / 0.47 ; 0.49 / 0.32 \\
ResViT 2D + 3D-refine              & tumour & 0.91 & 0.74 & 0.89 & 0.74 & 0.43 / 0.43 ; 0.50 / 0.37 \\
ResViT 2D + 3D-refine              & cavity & 0.14 & 0.05 & 0.09 & 0.07 & 0.33 / 0.47 ; 0.49 / 0.32 \\
\bottomrule
\end{tabular}%
}
\end{table}
\section{Discussion}
\label{sec:discussion}
This study provides a controlled comparative analysis of multiple deep learning strategies for intraoperative ioUS-to-MRI synthesis in neuro-oncology, spanning adversarial (GAN) models, transformer-based architectures, and diffusion-based approaches across four inference regimes. Our results reveal several important insights into the relative strengths, limitations, and practical utility of current synthesis architectures.
First, no single architectural family consistently outperformed all others across all evaluation criteria, underscoring the inherently multidimensional nature of medical image synthesis assessment. CGAN-based models, particularly when combined with hybrid 2D+3D refinement, achieved the highest global SSIM values (up to 0.814 for Pix2Pix-2D+3D-refine), whereas transformer-based ResViT variants produced the strongest PSNR performance (up to 15.41~dB for ResViT-2.5D). In contrast, ResViT-Full-3D and diffusion-based SynDiff models achieved the lowest LPIPS values (0.166 and 0.186, respectively), indicating superior perceptual similarity. Rather than reflecting minor quantitative differences, this divergence reveals that different metrics reward fundamentally different image properties and may favour distinct architectural behaviours.
Additional analysis suggests that some of these apparent advantages may be misleading when interpreted in isolation. The superiority of GAN-based models in global SSIM was substantially attenuated when evaluation was restricted to the lesion and a 5~mm perilesional margin: on the strict 0~mm lesion mask, the 2D+3D-refinement GAN variants retained their SSIM lead (e.g., SwinPix2Pix-2D+3D-refine 0.757), but once a 5~mm margin was included, the ResViT family overtook every GAN variant on SSIM (ResViT-2D+3D-refine 0.667) and dominated the LPIPS ranking (ResViT-2.5D 0.120), with no GAN variant appearing in the top-five LPIPS positions. This indicates that much of the apparent GAN structural advantage originates from anatomically homogeneous background regions rather than from surgically relevant tissue. The distinction is clinically important, as performance in non-informative background contributes little to the practical value of intraoperative synthesis. These findings challenge the common practice of relying on global SSIM or PSNR as primary indicators of synthesis quality in clinically oriented studies.
The downstream evaluation further strengthens this interpretation. Among all tested metrics, LPIPS showed the strongest association with downstream segmentation utility (Pearson $r=-0.66$ with NSD\textsubscript{2mm} utility ratio; $p<0.001$), suggesting that perceptual similarity better captures the preservation of task-relevant anatomical information than conventional voxel-wise measures. Notably, SSIM correlated in the opposite direction ($r=-0.64$ with NSD\textsubscript{2mm} utility; $p<0.001$), indicating that pixel-domain structural similarity does not predict---and may in fact inversely predict---the preservation of features relevant to downstream tasks. Consistently, no GAN-based configuration ranked among the top-performing models for downstream utility, despite their favourable SSIM performance. Instead, SynDiff-2.5D most effectively preserved the performance of segmentation models trained on real T2-weighted MRI (pooled $U_{\text{Dice}}=0.55$, $U_{\text{NSD}}=0.61$), followed closely by ResViT-Full-3D (0.52 / 0.59) and SynDiff-2D (0.51 / 0.60). This discrepancy strongly suggests that optimisation toward conventional fidelity metrics does not necessarily translate into clinically meaningful utility, and may in some cases favour models that reproduce background appearance more faithfully than surgically relevant anatomy.
The influence of inference dimensionality was also noteworthy. Although full volumetric 3D synthesis is intuitively appealing because of its ability to exploit complete spatial context, this strategy did not consistently outperform lower-dimensional alternatives in our experiments. One plausible explanation is the mismatch between model complexity and dataset scale. Full-3D architectures impose substantially greater computational and optimisation demands, often requiring reduced batch sizes and more extensive training data to achieve stable convergence. In contrast, 2D and 2.5D approaches offer greater sample efficiency and more stable optimisation, while hybrid 2D-to-3D refinement partially restores volumetric consistency at modest computational cost. In small-to-moderate neuroimaging cohorts such as ReMIND, this hybrid compromise may be more practical than end-to-end volumetric generation.
Architectural differences also revealed distinct synthesis behaviours. Transformer-based models demonstrated strong intensity-domain fidelity and favourable perceptual similarity, potentially reflecting their ability to capture broader contextual dependencies; this was most evident in ResViT's dominance of both the PSNR and LPIPS rankings, as well as its superior performance on the clinically relevant 5~mm lesion margin. Diffusion-based models similarly produced perceptually plausible outputs that translated more effectively into downstream task performance. In contrast, GAN-based architectures appeared comparatively better aligned with traditional structural similarity metrics but less effective at preserving information useful for subsequent clinical tasks---a divergence that was particularly pronounced for the resection cavity class, where ResViT-2D+3D-refine---the strongest model for tumour utility---achieved a cavity utility ratio of only 0.08, confirming that strong synthesis on one target does not guarantee clinically meaningful quality on another. Taken together, these findings suggest that architectural selection should be guided not by a single benchmark ranking but by the intended operational objective.
The multi-task synthesis experiments are particularly encouraging from a translational perspective. Simultaneous prediction of multiple MR sequences---specifically T2-weighted and FLAIR-like outputs---incurred only modest performance degradation relative to single-task models (at most 0.010 SSIM and 0.20~dB PSNR for the GAN family). Moreover, synthetic FLAIR retained downstream utility more consistently than synthetic T2w (73\,\% vs.\ 61\,\% of the real-MR lesion Dice ceiling), suggesting that the multi-task approach is operationally worthwhile despite marginally lower per-voxel fidelity. This indicates that shared representation learning across related sequences may be feasible without substantial loss of diagnostic or task-relevant fidelity.
The subgroup analysis revealed that synthesis quality was essentially independent of histological grade---no LGG versus HGG comparison reached significance on any metric---an encouraging property for clinical generalisation across tumour types. Reoperation status, however, lowered PSNR in the GAN family and ResViT-Full-3D (0.6--0.8~dB; ResViT-2.5D and the SynDiff variants were largely insensitive) and degraded perceptual quality (significant LPIPS effects in ResViT-2D and ResViT-2.5D), without detectably affecting SSIM---a useful warning that SSIM alone may mask clinically meaningful subgroup effects. From a downstream perspective, the reoperation effect on tumour utility was uniform and substantial (a drop of 0.15--0.34 in $U_{\text{Dice}}$ across all top configurations), whereas the cavity utility response was model-dependent: the SynDiff family recovered cavity utility better in reoperation cases, while the ResViT family was most adversely affected. These patterns add a clinically actionable layer to the recommendations derived from the main benchmark: the choice of synthesis architecture should be conditioned not only on the surgical phase but also on the patient's surgical history.
Our findings extend prior work in brain MRI/ioUS synthesis. Dorent et al.~\citep{dorent_mhvae_2023} demonstrated the feasibility of unified bidirectional MRI/ioUS synthesis on the ReMIND dataset using hierarchical multimodal variational modelling, establishing an important benchmark for this task. However, prior studies have generally focused on proposing individual architectures under fixed experimental conditions, limiting direct comparison across paradigms within a unified framework. Our study demonstrates that performance differences may depend substantially on inference strategy and practical optimisation constraints---such as dataset scale and available GPU memory---rather than on architectural innovation per se, and that the choice of evaluation metric profoundly influences which architecture appears superior.
At the same time, synthetic medical imaging must be interpreted cautiously. Improved perceptual realism does not guarantee anatomical trustworthiness. Generative models may smooth margins, alter pathology boundaries, or hallucinate plausible but incorrect structures---particularly in acoustically degraded regions where the input ioUS signal provides insufficient constraints. In surgical navigation contexts, even subtle inaccuracies may carry disproportionate clinical consequences, and the lower cavity utility ratios observed across all configurations ($U_{\text{Dice}}$ 0.08--0.42) highlight that the resection bed remains the most challenging and clinically sensitive region for current synthesis models.
From a broader perspective, our findings position synthesis as complementary---rather than necessarily superior---to direct multimodal registration. Classical MRI--ioUS registration methods such as cDRAMMS~\citep{machado_deformable_2019}, or modern optimisation-based frameworks, remain highly effective, particularly in pre-resection scenarios with preserved anatomy. However, registration directly across the MRI--ultrasound modality gap remains fundamentally challenging. Cross-modal synthesis offers an alternative strategy: rather than aligning incompatible modalities, one may reduce the domain gap before applying conventional downstream tools. Whether synthesis should serve primarily as a navigation aid, a preprocessing step for registration, or a bridge for MRI-trained AI models remains an open question that prospective studies should address.
\subsection{Limitations}
Several limitations must be acknowledged. First, the benchmark relies on a single institutional cohort (ReMIND), limiting external generalisability across centres and acquisition protocols. Second, the dataset remains modest in size for training data-intensive generative models, particularly full-3D transformers and diffusion architectures, which may explain the underperformance of Full-3D variants relative to their theoretical capacity. Third, all experiments assume approximate pre-alignment through rigid co-registration, meaning the benchmark evaluates synthesis under partially standardised geometry rather than raw unaligned multimodal translation. Fourth, downstream segmentation performance may partly reflect residual domain shift between synthetic and native MRI, in addition to the synthesis quality itself. Fifth, subgroup analyses such as the reoperation stratum remain exploratory and should be interpreted cautiously given the small cell sizes (as few as three patients per joint subgroup).
\subsection{Future Directions}
Future work should focus on larger multi-centre validation, deformable alignment-aware generative frameworks, and uncertainty-aware synthesis to expose unreliable regions---particularly in the resection cavity, where current models show the greatest limitation. Tighter integration between synthesis and downstream clinical tasks such as registration or intraoperative navigation is also warranted. Hybrid frameworks that jointly optimise synthesis and anatomical correspondence may be particularly promising. From an evaluation perspective, the strong correlation between LPIPS and downstream utility observed here argues for wider adoption of perceptual and task-based metrics as primary endpoints in future synthesis benchmarks, alongside or in preference to global SSIM. Ultimately, clinical adoption will depend not only on image realism but also on robustness across patient populations, interpretability of failure modes, and demonstrable benefit in real intraoperative workflows.
\section*{Acknowledgements}
The authors thank Mª Luz de Andrés Loste, documentalist librarian at Hospital Universitario Río Hortega, for her help with literature retrieval and reference verification, an increasingly valuable task in times of fabricated references.
\section*{Funding}
This research received no specific grant from any funding agency in the public, commercial or not-for-profit sectors.
\section*{Conflicts of Interest}
The authors declare no competing interests.
\section*{Data Availability}
The code supporting this study will be made available in a public repository upon acceptance of the manuscript. The data can be requested from the corresponding author.
\section*{Generative AI Disclosure}
The authors used Claude (Anthropic) to correct English usage and improve readability; all AI-assisted text was subsequently reviewed and verified by the authors, who take full responsibility for the content. PaperBanana was used to generate some of the schematic figures.

\end{document}